\documentclass[11pt]{article}

\usepackage[final]{acl}

\usepackage{times}
\usepackage{latexsym}
\usepackage{helvet}
\usepackage[T1]{fontenc}

\usepackage[utf8]{inputenc}

\usepackage{microtype}

\usepackage{inconsolata}

\usepackage{graphicx}

\usepackage[utf8]{inputenc} 
\usepackage[T1]{fontenc}    
\usepackage{url}            
\usepackage{booktabs}       
\usepackage{amsfonts}       
\usepackage{nicefrac}       
\usepackage{microtype}      
\usepackage[table]{xcolor}         
\usepackage[HTML]{xcolor}
\usepackage{graphicx}
\usepackage{caption}
\usepackage{booktabs}
\usepackage{amsmath}
\usepackage{amssymb}
\usepackage{pifont}
\usepackage{subcaption}
\usepackage{stfloats}
\usepackage{wrapfig}
\usepackage{multirow}
\usepackage{amssymb}
\usepackage{xcolor}
\usepackage{xspace}
\usepackage{dsfont}
\usepackage{enumitem}
\usepackage{wrapfig}  
\usepackage{xcolor}
\usepackage{arydshln}
\usepackage{tcolorbox}
\tcbuselibrary{skins, breakable, theorems}
\usepackage{soul}

\definecolor{TakeawayBg}{HTML}{F4F5FF} 
\definecolor{TakeawayLabel}{HTML}{666666} 
\definecolor{TakeawayText}{HTML}{000000}  

\sethlcolor{TakeawayBg}

\tcbset{
  takeaway/.style={
    enhanced,
    colback=TakeawayBg,
    colframe=TakeawayText,
    arc=2mm,
    boxrule=0.8pt,
    left=3mm,right=3mm,
    top=2mm,bottom=2mm,
    fonttitle=\bfseries,
    coltitle=white,
    attach boxed title to top left={
      yshift=-2mm,
      xshift=3mm
    },
    boxed title style={
      colback=TakeawayLabel,
      colframe=TakeawayLabel,
      arc=1mm,            
      left=2mm,right=2mm,
      top=0.1mm,bottom=0.1mm
    },
  }
}

\title{Look Light, Think Heavy: What Multimodal Chain-of-Thought Reasoning Can and Cannot Do}

\author{
\textbf{Zhuoran Jin}$^{1, 2}$, 
\textbf{Kejian Zhu}$^{1, 2}$,
\textbf{Hongbang Yuan}$^{1, 2}$,
\textbf{Yupu Hao}$^{1, 2}$,\\
\textbf{Pengfei Cao}$^{1, 2}$\textbf{,}
\textbf{Yubo Chen}$^{1, 2}$\textbf{,}
\textbf{Kang Liu}$^{1, 2}$\textbf{,}
\textbf{Jun Zhao}$^{1, 2, *}$\\
$^1$ The Laboratory of Cognition and Decision Intelligence for Complex Systems,\\
Institute of Automation, Chinese Academy of Sciences, Beijing, China \\
$^2$  School of Artificial Intelligence, University of Chinese Academy of Sciences, Beijing, China\\
\{zhuoran.jin, kliu, jzhao\}@nlpr.ia.ac.cn, zhukejian2025@ia.ac.cn}

\begin{document}

\maketitle

\begin{abstract}

Chain-of-Thought (CoT) has become a standard method for improving reasoning capabilities in large language models (LLMs) by eliciting step-by-step thinking, but its effectiveness in multimodal tasks remains unclear.
In this paper, we aim to systematically investigate the key question: \textit{What can multimodal Chain-of-Thought reasoning do, and where and why does it fall short?}
To this end, we evaluate 12 multimodal tasks across perception and reasoning categories using both 14 non-reasoning models and 8 reasoning models.
Our analysis reveals several important findings: (1) CoT is not a free lunch and should be used selectively depending on the specific requirements of each task.
For perception tasks, CoT can lead to undesirable side effects, such as reduced performance in visual grounding and object counting.
In contrast, it proves effective for reasoning tasks involving mathematical, scientific, and multi-image reasoning;
(2) Compared to original models, existing open-source multimodal reasoning models often yield only marginal overall improvements, possibly due to an overemphasis on mathematical reasoning at the expense of broader capabilities;
(3) Visual reasoning remains a key bottleneck for current multimodal CoT, as models exhibit a ``\textit{Look Light, Think Heavy}'' pattern where verbal reflection rises and falls during reasoning, whereas visual reflection consistently diminishes.
These findings suggest that while multimodal CoT handles verbal reflection relatively well, it lacks the ability to maintain deep visual introspection throughout the reasoning process.

\end{abstract}

\def\thefootnote{*}\footnotetext{Corresponding author.
}\def\thefootnote{\arabic{footnote}}

\section{Introduction}

Large language models (LLMs) \citep{gpt4, llama3, qwen25, anthropic2024claude3} such as OpenAI's o1 \cite{openai2024o1} and Deepseek-R1 \cite{deepseekr1}, which exhibit strong reasoning capabilities, have recently garnered significant attention.
These slow-thinking systems leverage Chain-of-Thought (CoT) reasoning \cite{CoT} during inference time, enabling deeper and longer reasoning and reflection processes and achieving advanced performance on complex tasks such as math and coding reasoning.
While recent research has made notable progress in textual reasoning, addressing real-world tasks such as interpreting scientific diagrams \cite{mmmu}, solving geometry problems \cite{mathvista}, and tackling visual puzzles \cite{visualpuzzles} continues to rely on incorporating visual information.

\begin{figure}[!th]
  \centering
  \begin{subfigure}[b]{0.43\textwidth}
    \includegraphics[width=\linewidth]{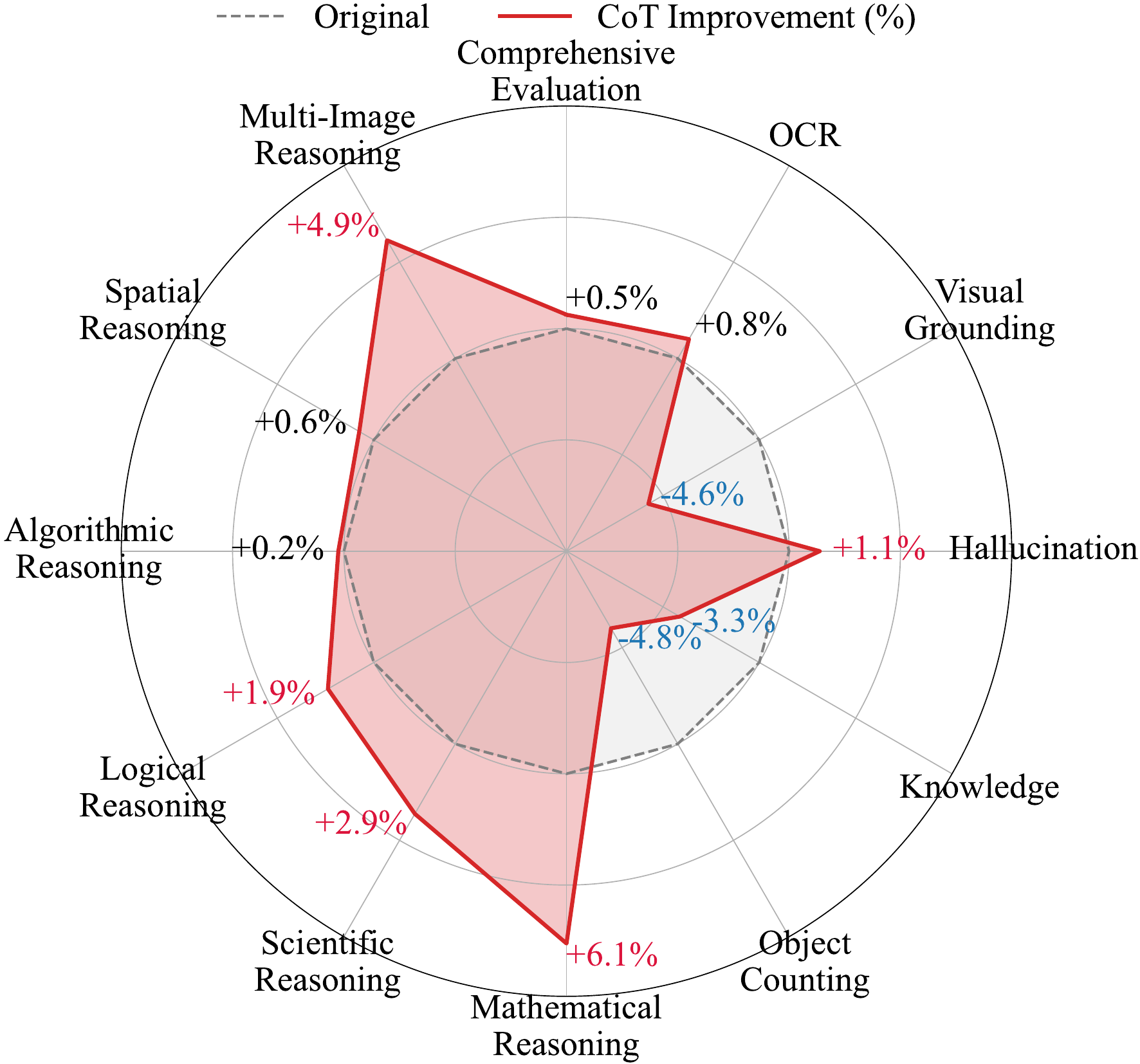}
    
    \caption{CoT vs.\ direct answer.}
      \label{fig:intro_1}
  \end{subfigure}
    \hfill
  \begin{subfigure}[b]{0.4\textwidth}
    \includegraphics[width=\linewidth]{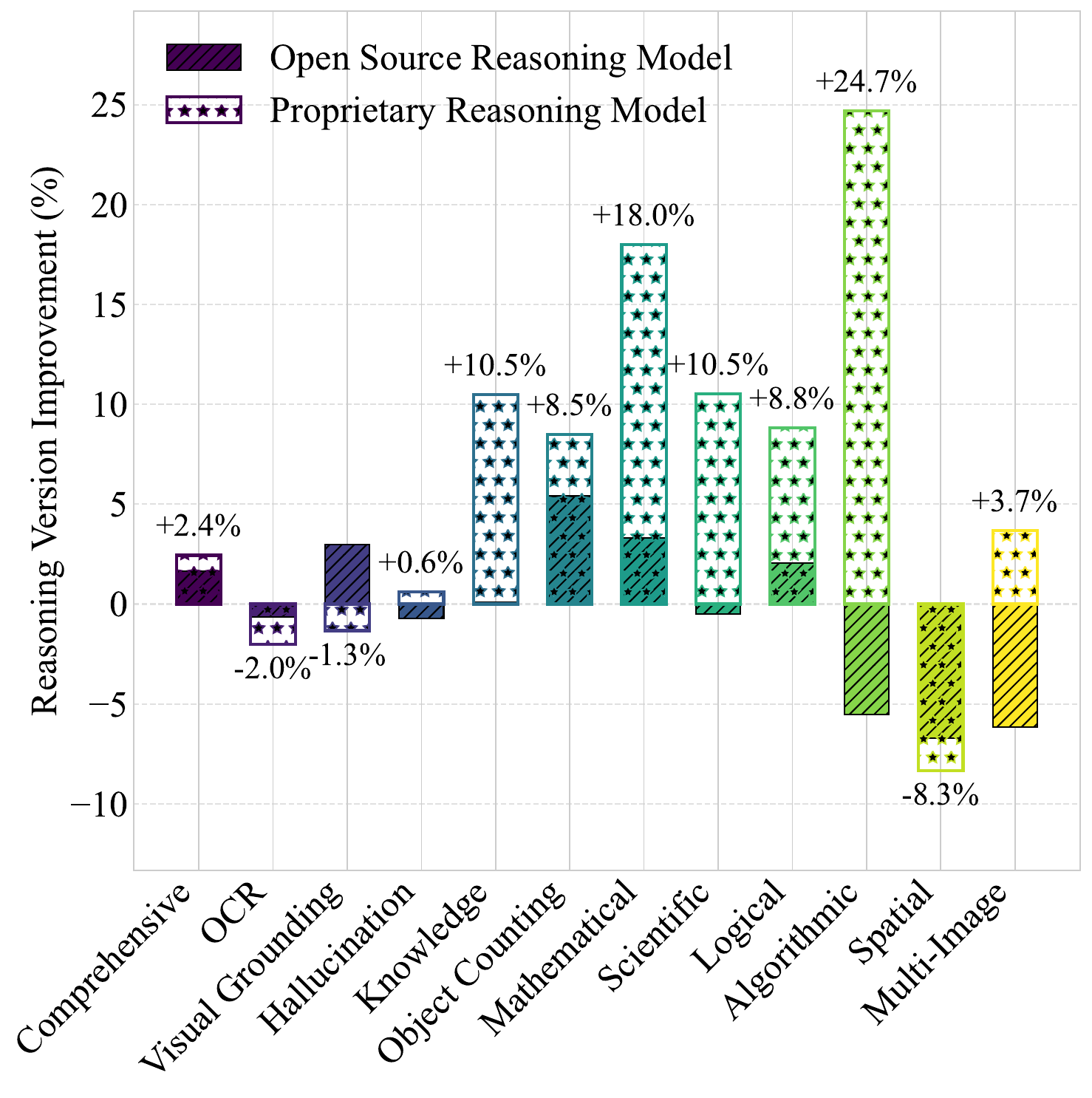}
    
    \caption{Reasoning model vs.\ base model. The \textit{Proprietary Reasoning Model} refers to Gemini-2.0-Flash-Thinking, while the \textit{Open-Source Reasoning Model} represents the average performance of the five models in Section~\ref{Comparison Between Non-Reasoning and Reasoning Models}.}
      \label{fig:intro_2}

  \end{subfigure}
  \hfill
  \begin{subfigure}[b]{0.5\textwidth}
    \includegraphics[width=\linewidth]{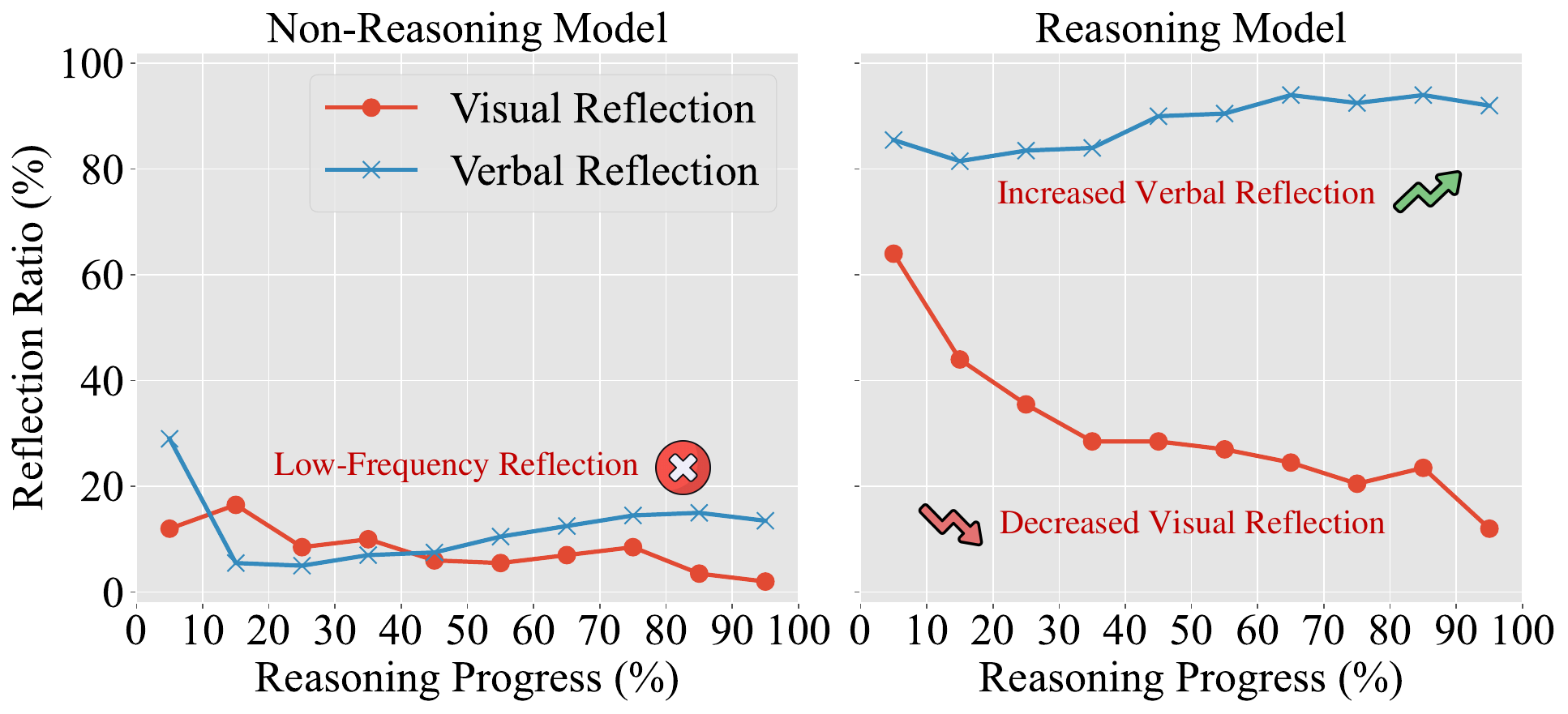}
  \caption{``\textit{Look Light, Think Heavy}'' pattern in multimodal CoT. The reasoning process indicates the first $x\%$ of the CoT.}
      \label{fig:intro_3}

  \end{subfigure}

  \caption{Main findings of multimodal CoT reasoning.}
  \label{fig:intro}
\end{figure}

Recently, an increasing number of studies have explored unlocking the CoT reasoning capabilities of multimodal large language models (MLLMs) \citep{openai2024gpt4o, deepmind2025geminiflash, qwen25vl}.
Similar to textual reasoning \cite{tocot}, multimodal CoT has been predominantly explored in the context of mathematical reasoning, with evaluations commonly conducted on benchmarks such as MathVista \cite{mathvista}, MathVerse \cite{MathVerse} and MATH-Vision \cite{MATH-Vision}.
However, the scope of multimodal tasks extends well beyond mathematical reasoning.
Given that CoT reasoning introduces additional inference overhead and complexity, it remains an open question whether CoT can consistently improve performance across a broad range of multimodal tasks.

In this paper, we aim to systematically investigate the key question: \textit{What can multimodal Chain-of-Thought reasoning do, and where and why does it fall short?}
First, we categorize multimodal tasks along two dimensions: \textbf{multimodal perception} and \textbf{multimodal reasoning}.
Multimodal perception tasks include comprehensive evaluation, OCR, visual grounding, hallucination, knowledge and object counting, while multimodal reasoning tasks include mathematical, scientific, logical, algorithmic, spatial and multi-image reasoning.
Then, we conduct experiments with 14 \textbf{non-reasoning models} (\textit{e.g.}, Qwen2.5-VL \cite{qwen25vl}, Gemma-3 \cite{gemma3}, GPT-4.1 \cite{openai2025gpt41}) and 8 \textbf{reasoning models} (\textit{e.g.}, QVQ \cite{QVQ}, Skywork-R1V2 \cite{SkyworkR1V2}, Gemini-Thinking \cite{google2025gemini2.5}), to evaluate the strengths and pitfalls of multimodal CoT.
Finally, we investigate the limitations of current multimodal CoT reasoning by exploring its \textbf{external behaviours} and \textbf{internal mechanisms}.
Based on the above analytical framework, we further decompose the central issue of multimodal CoT reasoning into three research questions (RQs).

\textbf{RQ1: What multimodal CoT can and cannot do for both perception and reasoning tasks?} We compare the performance of direct answering and CoT reasoning across 12 multimodal perception and reasoning tasks.
We find that \textbf{CoT is not a free lunch and should be used selectively depending on the specific requirements of each task}.
As shown in Figure \ref{fig:intro_1}, CoT can lead to undesirable side effects in perception tasks such as \textbf{visual grounding}, \textbf{knowledge-based VQA}, and \textbf{object counting}.
For reasoning tasks, CoT proves particularly effective in domains such as \textbf{mathematical}, \textbf{scientific}, and \textbf{multi-image reasoning}, where it consistently improves performance across almost all models. For \textbf{logical} and \textbf{algorithmic reasoning}, the effectiveness of CoT is model-dependent. Larger models tend to benefit from CoT, whereas smaller models often experience negative gains.

\textbf{RQ2: Can multimodal reasoning models outperform base models through test-time scaling?} 
Although reinforcement learning with verified rewards (RLVR) has shown great potential in LLMs, enabling them to generate longer CoT with emergent reflective abilities \citep{qwq32b, DBLP:journals/corr/abs-2501-12599, dapo}, it remains unclear whether the same strategy can be effectively extended to MLLMs.
We compare non-reasoning models with their reasoning variants.
As shown in Figure \ref{fig:intro_2}, we reveal that \textbf{existing open-source multimodal reasoning models often achieve only marginal improvements in average performance across a wide range of tasks}.
This may be attributed to their predominant training on mathematical problems using RLVR, which tends to \textbf{overemphasize mathematical reasoning while neglecting broader reasoning capabilities}. 
In contrast, commercial reasoning models such as Gemini-2.0-Flash-Thinking demonstrate substantial and consistent gains across diverse reasoning tasks.

\textbf{RQ3: What are the key limitations that hinder the effectiveness of multimodal CoT?} Building on the above analysis, we observe that current multimodal CoT still faces several challenges.
First, we design a set of \textcolor[HTML]{d62728}{visual} and \textcolor[HTML]{1f77b4}{textual} reasoning probes based on several multimodal reasoning tasks.
Our findings indicate that \textbf{visual reasoning is critical to the effectiveness of multimodal CoT and currently constitutes a primary bottleneck limiting its overall performance}.
Subsequently, we decompose reflective behaviours in multimodal CoT into \textcolor[HTML]{d62728}{visual reflection} and \textcolor[HTML]{1f77b4}{verbal reflection}.
As shown in Figure \ref{fig:intro_3}, we observe that existing multimodal reasoning models exhibit a ``\textit{Look Light, Think Heavy}'' pattern: \textbf{\textcolor[HTML]{1f77b4}{verbal reflection} follows a rise-and-fall trajectory, peaking in the middle of the verbal reasoning process, while \textcolor[HTML]{d62728}{visual reflection} steadily declines over time}.
Meanwhile, we also identify a persistent \textit{attention bias} in multimodal long CoT.
\textbf{During extended reasoning, models tend to allocate disproportionate attention to reasoning tokens while progressively neglecting visual tokens}.
These phenomena confirm that current multimodal CoT is more adept at verbal reflection during the reasoning process, yet struggles to maintain deep visual introspection.

We further discuss future directions for advancing multimodal CoT reasoning.
We observe that when critical visual information is missing, current models are unable to reflect on the visual input and abstain from answering accordingly.
This underscores the necessity for MLLMs to possess visual introspection capabilities.
Moreover, to address the visual bottlenecks of current models, they should be equipped with mechanisms to leverage external tools that enhance visual understanding.
Recent advancements, such as the \textit{think-with-image} paradigm adopted by OpenAI's o3 and o4 \cite{openai2024o3o4mini}, may represent a promising direction.

\section{Problem Formulation}
\label{Problem Formulation}

\subsection{Multimodal Chain-of-Thought}

Given a set of one or more image inputs $I$, a textual question $q$, and a CoT prompting prefix $p_{\text{c}}$, the model $M$ generates an output sequence as follows: $r, a = M(I, p_{\text{c}}, q)$.
Here, $r$ denotes a long CoT sequence that captures the step-by-step reasoning process leading to the final answer $a$.
The prompt $p_{\text{c}}$ can be \textit{``Please first think about the reasoning process as an internal monologue and then provide the final answer.''}.
In contrast, direct answering without CoT yields a shorter output sequence containing only the final answer: $a = M(I, p_{\text{d}}, q)$, $p_{\text{d}}$ can be \textit{``Please generate the answer directly.''}.

\subsection{Perception and Reasoning Tasks}

To holistically evaluate the impact of CoT, we categorize multimodal tasks along two dimensions: multimodal \textbf{perception} and \textbf{reasoning}.  
The perception category includes comprehensive evaluation, OCR, visual grounding, hallucination detection, knowledge-based VQA, and object counting, which focus on fine-grained visual understanding and cross-modal alignment.
The reasoning category includes mathematical, scientific, logical, algorithmic, spatial, and multi-image reasoning, which emphasize multi-step reasoning grounded in both visual and textual inputs.
The detailed descriptions of 12 tasks are in Appendix \ref{Task Details}.

\subsection{Evaluation Models}

We conduct experiments on both \textbf{non-reasoning} (general) and \textbf{reasoning} models.
Compared with non-reasoning models, reasoning models are capable of generating much longer CoT sequences and exhibit a certain degree of reflection, enabling them to perform self-correction in CoTs.
For non-reasoning models, we compare their performance under direct answering and CoT. For reasoning models with test-time scaling, we analyze performance differences with their corresponding non-reasoning models.
For details on the models and prompts used, please refer to Appendix \ref{Prompts}.

\section{Strengths and Pitfalls of Multimodal Chain-of-Thought}

In this section, we conduct a thorough analysis of the strengths and pitfalls of CoT reasoning in MLLMs.
We first compare the performance of CoT with direct answering across perception and reasoning tasks.
We then examine the differences between non-reasoning and reasoning models.

\subsection{Comparison Between Direct Answer and Chain-of-Thought}

\begin{figure*}[t]
    \centering
    \includegraphics[width=1.0\linewidth]{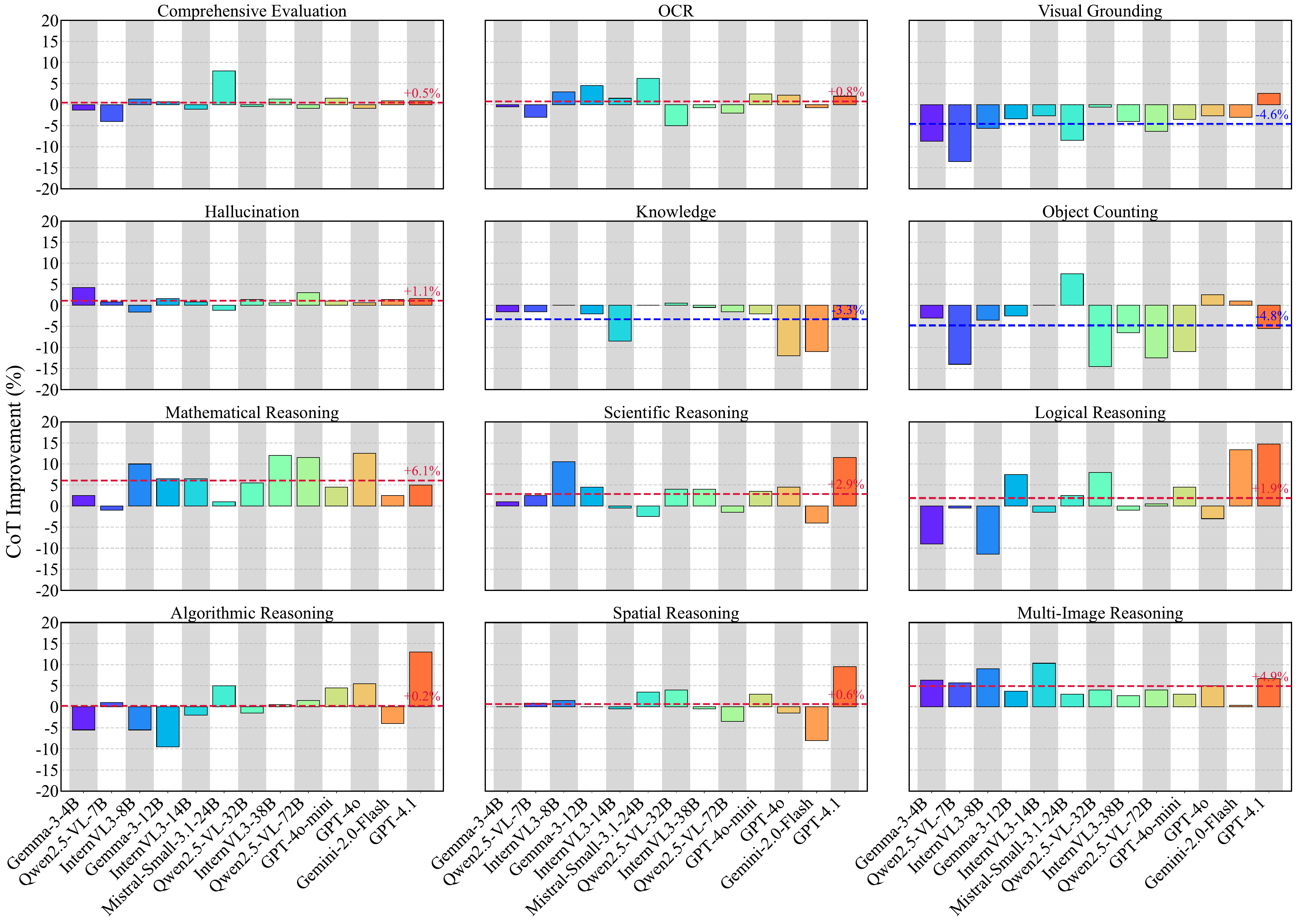}
    \caption{Comparison between direct answer and CoT. Y-axis shows the performance gain of CoT.}

    \label{fig:CoT}
\end{figure*}

To understand the strengths and limitations of CoT, we first compare its performance with direct answering across a range of multimodal perception and reasoning tasks.
As illustrated in Figure~\ref{fig:CoT}, the effectiveness of CoT is inconsistent across different types of multimodal tasks.
For perception tasks, it may lead to marginal or even negative effects. In particular, we observe average performance drops of \textbf{4.6\%}, \textbf{3.3\%}, and \textbf{4.8\%} on visual grounding, knowledge-based VQA, and object counting, respectively.
This degradation may be attributed to the fact that CoT introduces additional reasoning steps that are unnecessary or even distracting for perception-oriented tasks.

\vspace{-5pt}
\begin{tcolorbox}[takeaway,title={Takeaway 3.1.1 for RQ1}]
CoT is not a free lunch and should be applied selectively according to the type of multimodal task.
It is generally more effective for reasoning tasks but may introduce side effects in perception tasks such as visual grounding, knowledge VQA, and object counting.
\end{tcolorbox}
\vspace{-5pt}

In contrast to perception tasks, CoT is more effective in reasoning tasks. We observe performance improvements of \textbf{6.1\%}, \textbf{2.9\%}, and \textbf{4.9\%} on mathematical, scientific, and multi-image reasoning tasks, respectively.
For mathematical and scientific reasoning, MLLMs demonstrate similar improvements to those observed in LLMs, as these tasks primarily depend on text-dominant reasoning following basic visual understanding.
For multi-image reasoning tasks, we observe that MLLMs tend to describe each image in CoT, and subsequently perform reasoning based on the aggregated textual descriptions.

For logical and algorithmic reasoning, which rely more heavily on reasoning over visual information, we find that the effectiveness of CoT is closely related to model scale.
Larger models benefit from CoT reasoning, while smaller models often show limited or even degraded performance.

\vspace{-5pt}
\begin{tcolorbox}[takeaway,title={Takeaway 3.1.2 for RQ1}]
CoT is more effective for problems involving mathematical, scientific, and multi-image reasoning, where it consistently improves performance across almost all models.
For logical and algorithmic reasoning, the effectiveness of CoT varies with model scale. While larger models often benefit from CoT, smaller models tend to experience negative gains.
\end{tcolorbox}
\vspace{-5pt}

\begin{figure*}
    \centering
    \includegraphics[width=1.0\linewidth]{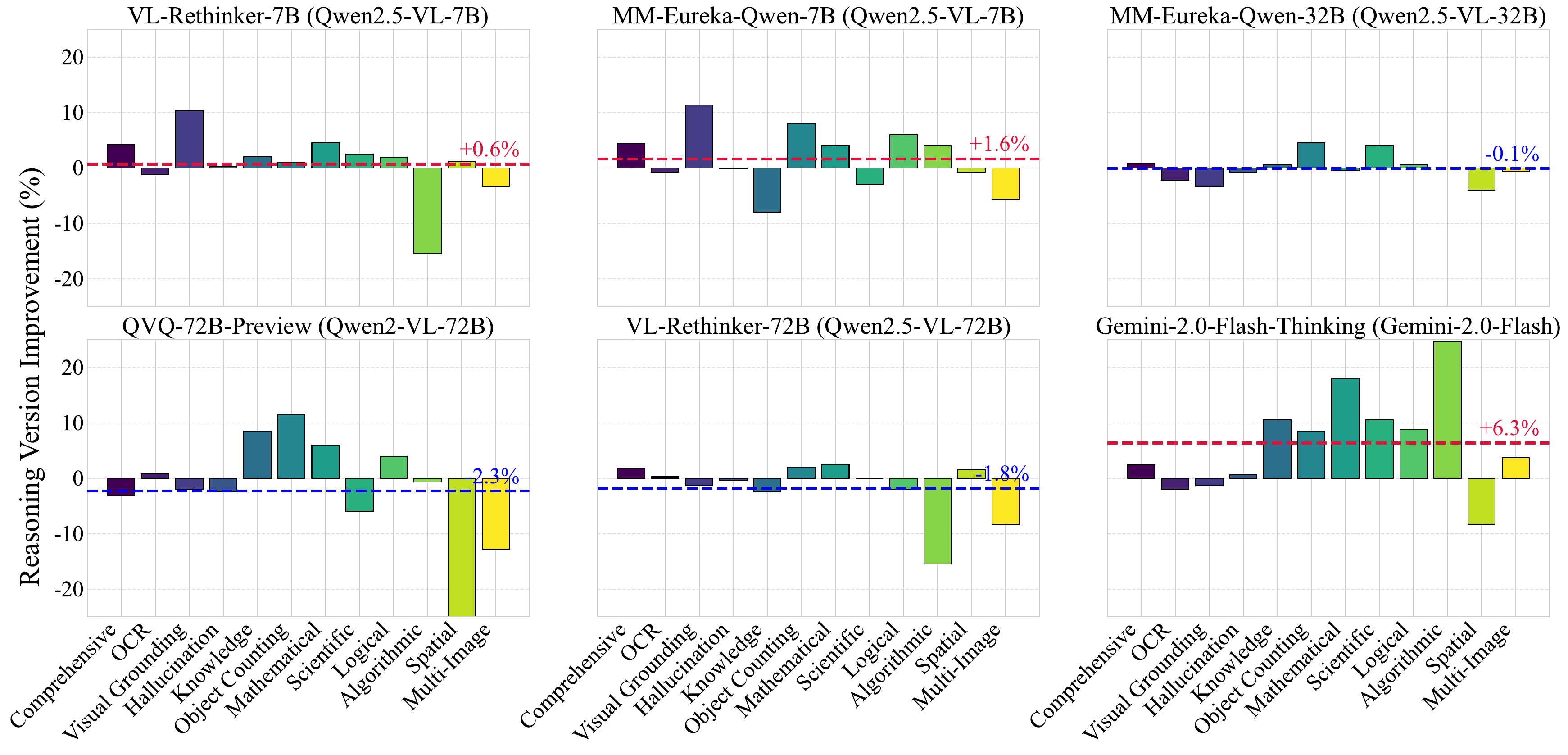}
    \caption{Comparison between non-reasoning models and reasoning models.}
    \label{fig:thinking}
\end{figure*}

\subsection{Comparison Between Non-Reasoning and Reasoning Models}
\label{Comparison Between Non-Reasoning and Reasoning Models}

Many studies \citep{deepseekr1, qwq32b, DBLP:journals/corr/abs-2501-12599, dapo} have trained original models into reasoning models using RLVR, enabling longer CoT and emergent reflection.
Although recent works have attempted to extend this strategy to MLLMs, it remains unclear whether it yields comparable multimodal reasoning abilities.
We compare five open-source models and one commercial model, evaluating both their original and reasoning-enhanced versions.

As shown in Figure~\ref{fig:thinking}, open-source multimodal reasoning models often exhibit only limited performance gains across diverse tasks.
One possible explanation is that these models are primarily trained on math-related questions with verified rewards, which leads to an overemphasis on mathematical reasoning while neglecting other reasoning abilities.
In contrast, Gemini-2.0-Flash-Thinking, as a commercial reasoning model, demonstrates substantial and consistent gains across diverse reasoning tasks, with a notable improvement of 24.7\% in algorithmic reasoning.
These observations highlight the need for new training paradigms that better generalize across various types of multimodal reasoning.

\vspace{-5pt}

\begin{tcolorbox}[takeaway,title={Takeaway 3.2.1 for RQ2}]
Existing open-source multimodal reasoning models show modest gains across a range of tasks. This may be attributed to their predominant training on mathematical problems using RLVR, which tends to prioritize mathematical reasoning while overlooking broader reasoning capabilities. 
It may be necessary to explore novel training paradigms for MLLMs.
\end{tcolorbox}
\vspace{-5pt}

\section{Shallow Visual Reflection in Multimodal Chain-of-Thought}

In this section, we experimentally investigate the role and significance of visual information analysis and reasoning in multimodal CoT generation.
Furthermore, we examine whether current multimodal reasoning models exhibit similar paradigms and limitations in their reasoning over visual information, based on both external reflection behaviours and internal attention mechanisms.

\subsection{Visual Reasoning Bottleneck in Multimodal Reasoning}
\label{sec:bottleneck}

\begin{figure}[t] 
    \centering
    \includegraphics[width=0.95\linewidth]{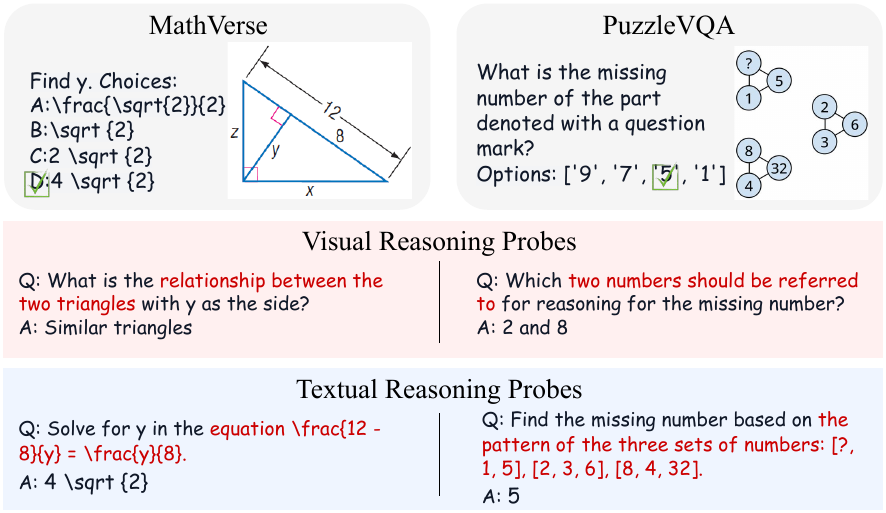}
    \caption{Examples of \textcolor[HTML]{d62728}{visual} and \textcolor[HTML]{1f77b4}{textual} reasoning probes for mathematics and logical reasoning tasks.}
    \label{fig:probe_samples}
\end{figure}

\begin{figure*}[thbp]
  \centering
  \begin{subfigure}[b]{0.88\textwidth} 
    \includegraphics[width=\linewidth]{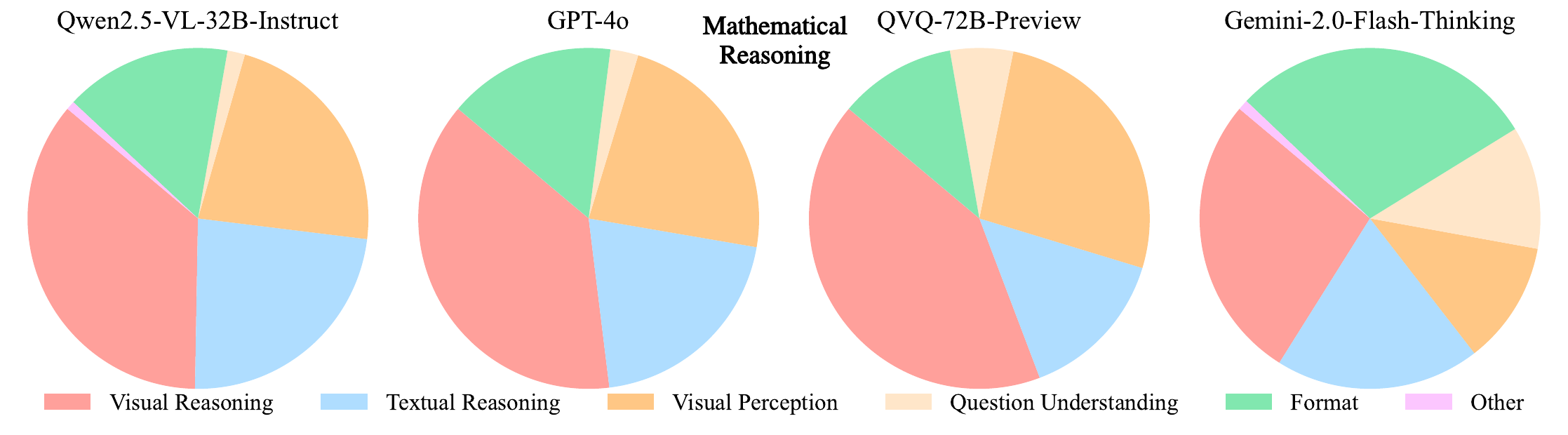}
  \end{subfigure}

  \vspace{-1ex} 

  \begin{subfigure}[b]{0.88\textwidth}
    \includegraphics[width=\linewidth]{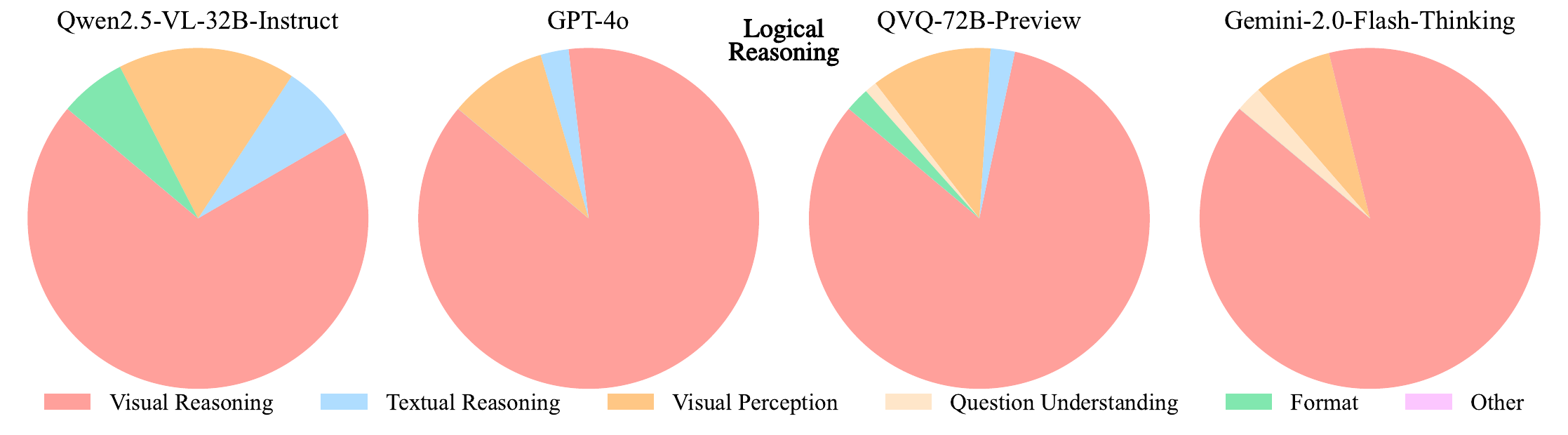}
  \end{subfigure}
  
  \caption {Error analysis of CoT in mathematical and logical reasoning.}
    \label{fig:error}

\end{figure*}

To investigate the role of visual reasoning in multimodal CoT, we first analyze CoT failure cases.
We provide detailed descriptions of error types in Appendix \ref{Implementation Details}.
As shown in Figure~\ref{fig:error}, a large proportion of errors arise from visual reasoning failures, particularly in logical reasoning tasks, where over 80\% are due to incorrect reasoning over visual information.
Then, we analyze the relative contributions of visual and textual reasoning to the overall solution process in multimodal reasoning tasks.
To this end, we design two types of reasoning probes: \textcolor[HTML]{d62728}{\textbf{visual reasoning}} and \textcolor[HTML]{1f77b4}{\textbf{textual reasoning}}. As illustrated in Figure \ref{fig:probe_samples}, \textcolor[HTML]{d62728}{visual reasoning} probes focus on subtasks of original problem that require analyzing and reasoning over visual information, such as identifying geometric similarity or detecting visual patterns. \textcolor[HTML]{1f77b4}{Textual reasoning} probes involve subtasks that rely only on reasoning which is independent of visual information, such as computing equations derived from visual analysis or identifying patterns within numerical sets. 
Importantly, both types of probes correspond to intermediate steps within the original multimodal reasoning tasks, contributing to the understanding of which parts of the solution process pose the greatest challenge for the model.

\begin{figure}[t] 
\centering
  \includegraphics[width=0.9\linewidth]{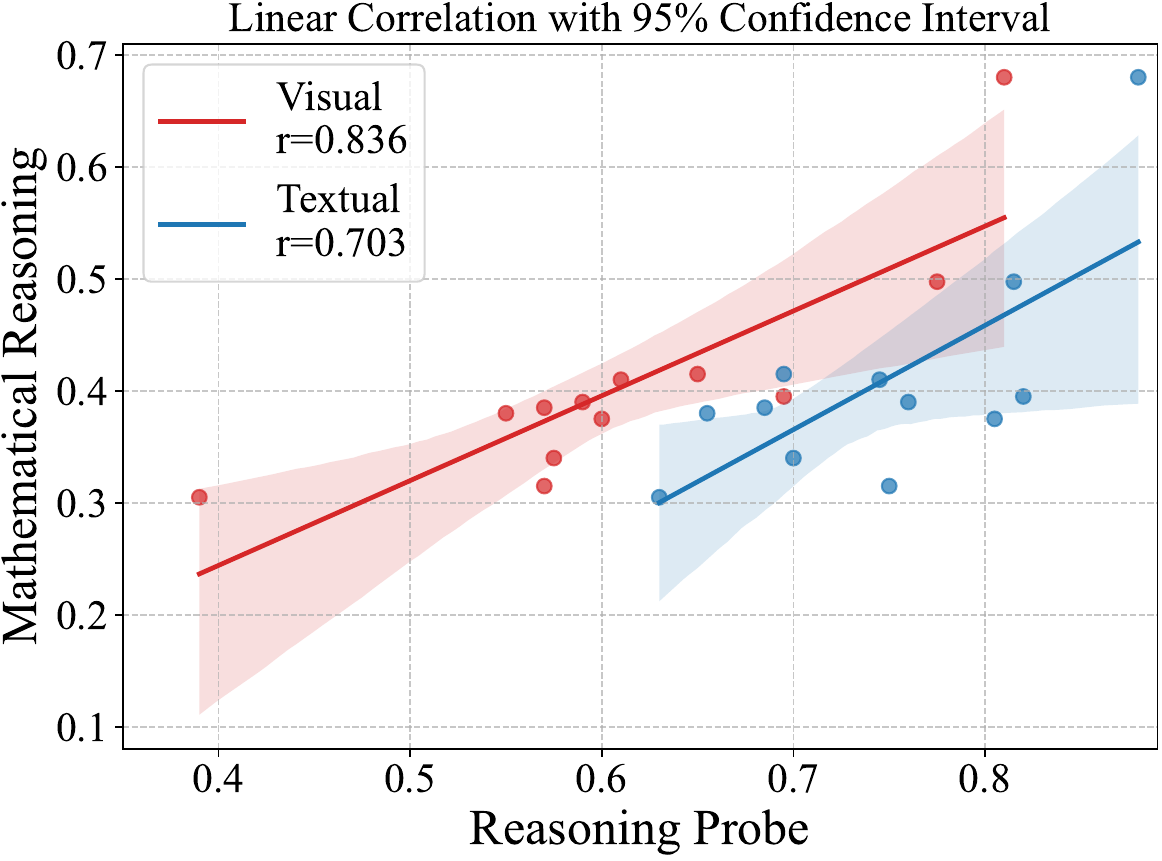} \hfill
  
  \caption {Correlation between overall task performance and reasoning probe accuracy of mathematical task across different models. \textcolor[HTML]{d62728}{Red} and \textcolor[HTML]{1f77b4}{blue} indicate visual reasoning and textual reasoning probes, respectively. r denotes the Pearson correlation coefficient. Additional results are in Figure \ref{fig:combined_correlation_ref_regression}.}
  
    \label{fig:combined_correlation_ref}
\end{figure}

We use o4-mini, which performs well on multimodal reasoning, to construct probe tasks. The correctness and suitability of the probes are verified with GPT-4.1, checking accuracy, uniqueness, and alignment with probe categories. Full prompt examples are provided in Appendix \ref{Reasoning Prompts}. We also conducted manual verification of the probe tasks. Mathematical probes achieved 93.0\% accuracy, and logical probes 88.5\%, indicating reliability.
We then evaluate general and reasoning models on these tasks and analyze the correlation between probe accuracy and original task performance. As shown in Figure~\ref{fig:combined_correlation_ref} and~\ref{fig:combined_correlation_ref_regression}, models consistently perform better on \textcolor[HTML]{1f77b4}{textual reasoning} than \textcolor[HTML]{d62728}{visual reasoning} probes, with an average gap of~20\%, highlighting the greater challenge of visual reasoning.

Furthermore, model performance on the original tasks shows a stronger correlation with performance on the visual reasoning probe, with Pearson correlation coefficients $r$ exceeding those for the textual probe in both tasks. These results suggest that visual reasoning remains a key challenge in current multimodal reasoning tasks and represents a bottleneck for current MLLMs. The strong correlation further underscores the critical role of visual reasoning in solving these tasks.

\begin{tcolorbox}[takeaway,title={Takeaway 4.1.1 for RQ3}]
Compared with textual reasoning probes, models show a 20\% performance drop on visual ones.
Moreover, visual probe accuracy shows a stronger correlation with overall task performance, highlighting that visual reasoning remains a key bottleneck of MLLMs.
\end{tcolorbox}

\subsection{Reflection Behaviours in Multimodal Chain-of-Thought}
\label{sec:analysis_reflect}

\begin{figure}[t]
    \centering
    \includegraphics[width=1.0\linewidth]{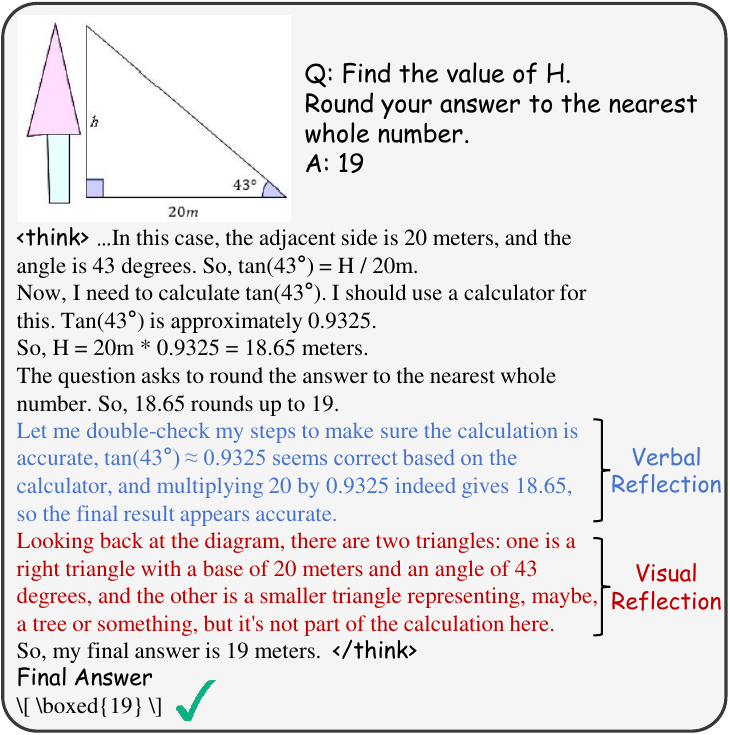}
    \caption{\textcolor[HTML]{d62728}{Visual reflection} and \textcolor[HTML]{1f77b4}{verbal reflection} behaviours in multimodal CoT.}
    \label{fig:reflection}
\end{figure}

\begin{figure*}[!t]
  \centering
  \begin{subfigure}[b]{0.95\textwidth} 
    \includegraphics[width=\linewidth]{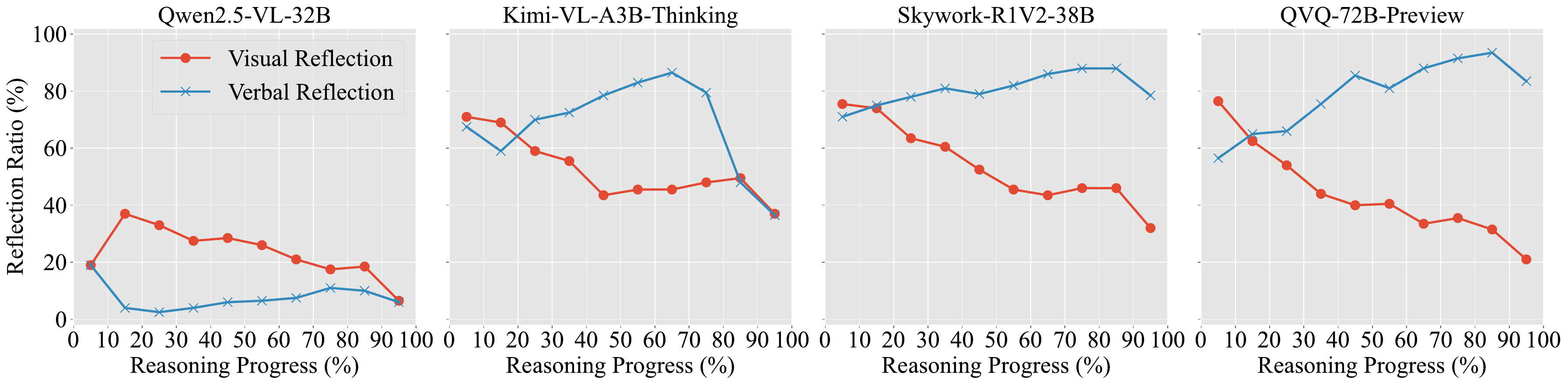}
  \end{subfigure}




  \begin{subfigure}[b]{0.95\textwidth}
    \includegraphics[width=\linewidth]{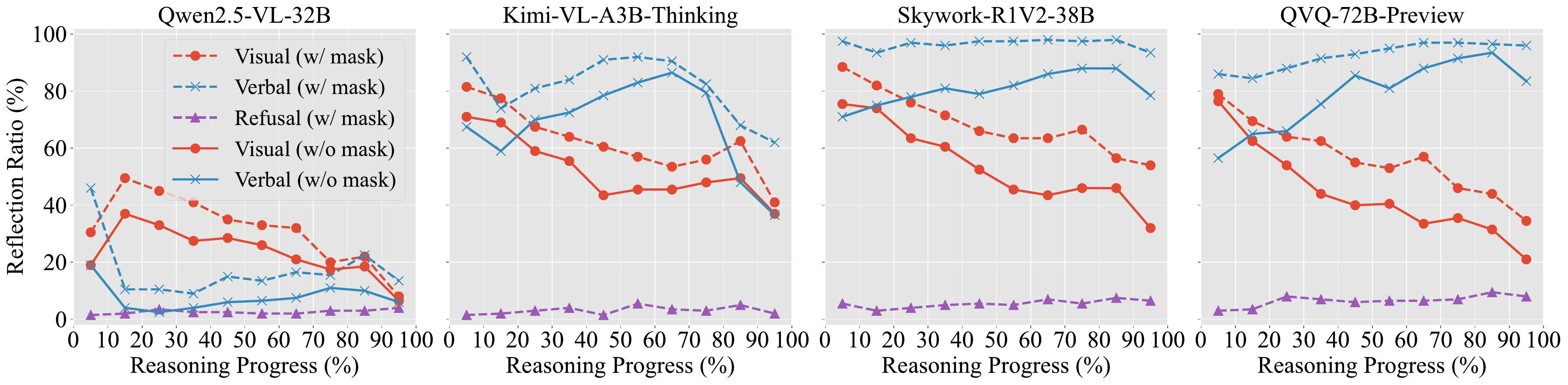}
  \end{subfigure}



\caption{Step-wise distribution of \textcolor[HTML]{d62728}{visual} and \textcolor[HTML]{1f77b4}{verbal} reflection in CoT.  
The two rows show MathVista and MathVista with missing critical visual information. More results are provided in Figure \ref{fig:reflection distribution2}.} 

  \label{fig:reflection distribution}
\end{figure*}

Given that visual reasoning is a primary limitation in multimodal CoT, we further examine what factors constrain models' ability to reason over visual information.
As reflection and self-verification are critical capabilities of reasoning models \citep{deepseekr1, openai2024o1}, with the potential to effectively improve reasoning accuracy, we examine whether such behaviours are exhibited in the CoT generated by current MLLMs.
For multimodal CoT, we categorize reflective behaviours into two types: \textcolor[HTML]{d62728}{\textbf{visual reflection}} and \textcolor[HTML]{1f77b4}{\textbf{verbal reflection}}.
As shown in Figure~\ref{fig:reflection}, \textcolor[HTML]{d62728}{visual reflection} refers to the model’s act of reconsidering its visual perception or interpretation.
This includes behaviours such as expressing uncertainty, doubt, or re-evaluating visual information, as illustrated by phrases like ``\textit{Let me double-check the image}'' or ``\textit{Maybe I misinterpreted the object in the picture}''.
\textcolor[HTML]{1f77b4}{Verbal reflection}, in contrast, refers to the model’s introspection on its own reasoning process.
This involves the model recognizing, questioning, or revising its intermediate reasoning steps or final conclusions, as illustrated by phrases such as ``\textit{Wait, my earlier assumption might be wrong}'' or ``\textit{This line of reasoning may not be sufficient}''.
We divide each CoT sequence into ten equal-length segments based on token count and use GPT-4.1 to annotate the presence of reflective behaviours at each step, using the prompt shown in Table~\ref{appendix:prompt_verbal_visual_reflection}.

As shown in Figures~\ref{fig:reflection distribution} and \ref{fig:reflection distribution2}, reasoning models, such as QVQ-72B-Preview, exhibit noticeably more visual and verbal reflection behaviours compared to non-reasoning models like Qwen2.5-VL-32B, indicating a stronger tendency to actively verify the reliability of visual inputs and assess the soundness of their own reasoning processes.
However, we can observe that visual and verbal reflection follow opposite trends throughout the CoT.
While \textcolor[HTML]{1f77b4}{verbal reflection} increases and peaks mid-way through the reasoning process, \textcolor[HTML]{d62728}{visual reflection} diminishes over time, indicating that models tend to deepen their textual reasoning while progressively overlooking visual information. 

\begin{tcolorbox}[takeaway,title={Takeaway 4.2.1 for RQ3}]
Existing multimodal reasoning models exhibit a ``\textit{Look Light, Think Heavy}'' pattern, where \textcolor[HTML]{1f77b4}{verbal reflection} behaviours follow a rise-and-fall trend, peaking in the middle of the reasoning process. In contrast, \textcolor[HTML]{d62728}{visual reflection} declines steadily over time, suggesting a lack of deep visual introspection capabilities in current multimodal models.
\end{tcolorbox}

These findings reveal a key limitation of current multimodal CoT reasoning: \textbf{shallow \textcolor[HTML]{d62728}{visual reflection} contrasted with deep \textcolor[HTML]{1f77b4}{verbal reflection}}.
To further validate this observation, we deliberately occlude critical visual information in the images using mosaics and assess whether models demonstrate visual reflection behaviours that result in abstention from answering.
We find that when confronted with missing visual cues, current multimodal reasoning models exhibit a noticeable increase in both visual and verbal reflective behaviours.
However, despite engaging in such reflection, they show a limited ability to abstain from answering when appropriate, suggesting that current forms of visual reflection are shallow and fail to support reliable abstention when key visual information is missing.

\begin{tcolorbox}[takeaway,title={Takeaway 4.2.2 for RQ3}]
When confronted with missing critical visual information, current multimodal reasoning models exhibit an increase in both visual and verbal reflective behaviours. However, they show limited ability to abstain from answering despite engaging in such reflection.
\end{tcolorbox}

\subsection{Attention Bias in Multimodal Chain-of-Thought}
\label{sec:attention_bias}
\begin{figure}[t]
  \centering

  \begin{subfigure}[b]{0.4\textwidth}
    \includegraphics[width=\linewidth]{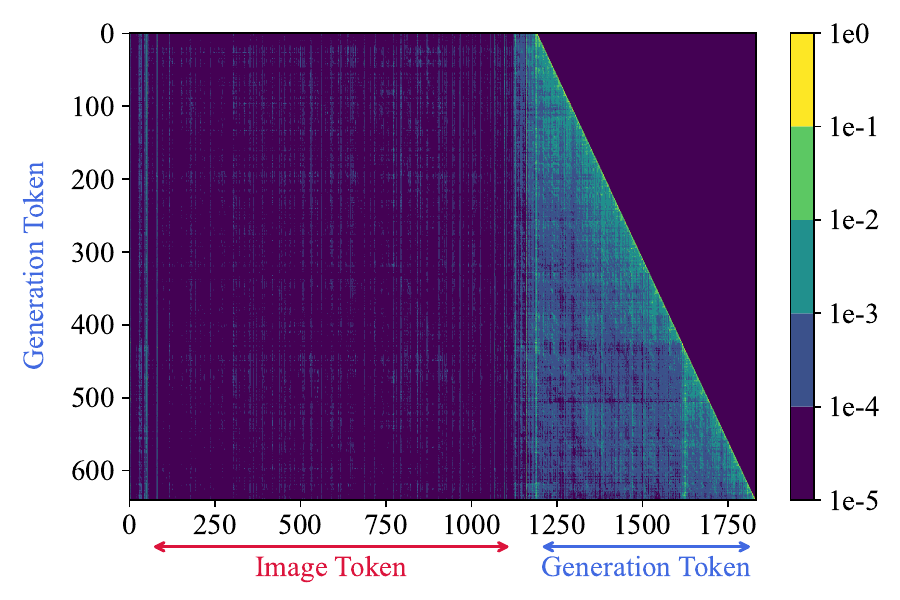}
  \end{subfigure}

  \begin{subfigure}[b]{0.4\textwidth}
    \includegraphics[width=\linewidth]{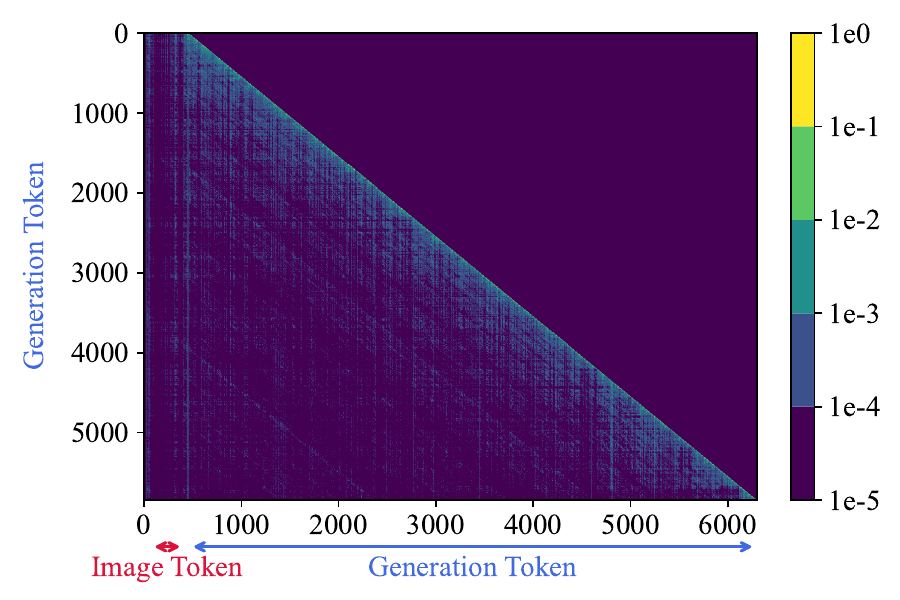}
  \end{subfigure}

\caption{Attention visualizations of Kimi-VL-A3B-Thinking on mathematical and logical reasoning, illustrating the cross-attention weights between the current token and the preceding tokens.}
  \label{fig:attention}
\end{figure}

To further investigate the underlying mechanism behind the observed shallow visual reflection, we analyze the internal attention patterns of the multimodal reasoning models during CoT generation.
We select Kimi-VL-A3B-Thinking as the representative reasoning model, with results for three additional models provided in Appendix \ref{appendix:additional_results}.
We prompt it to generate long-form CoT on mathematical and logical reasoning tasks, and subsequently visualize its internal attention weights to examine how attention is allocated throughout the reasoning process.
As shown in Figure~\ref{fig:attention}, during CoT generation the model exhibits a pronounced attention imbalance, increasingly prioritizing reasoning tokens while gradually neglecting visual inputs.
This attention bias may constrain the model’s ability to engage in effective visual reflection, leading it to over-rely on verbal reflection.

\begin{tcolorbox}[takeaway,title={Takeaway 4.3.1 for RQ3}]
During the generation of long CoT, multimodal reasoning models tend to allocate disproportionate attention weights to reasoning tokens while neglecting visual tokens. This phenomenon of attention bias not only shifts the model’s focus away from the original visual input but also potentially constrains its capacity for visual reflection.
\end{tcolorbox}

\section{Relate Works}

\textbf{Chain-of-Thought}.
CoT prompting improves performance on math and coding tasks by explicitly introducing intermediate reasoning steps~\cite{CoT, Self-Consistency, kojima2022large, zhou2022least, jin-etal-2024-tug, jin-etal-2024-cutting}. Recent studies \citep{s1, limo, yeo2025demystifying} explore test-time scaling strategies that generate longer CoT with reflection, promoting deeper reasoning. Besides, several works have extended CoT to multimodal tasks~\cite{zhang2023multimodal, mitra2024compositional, DBLP:conf/nips/HuSFROZSK24, DBLP:conf/acl/HeLBHTSHHHZLQL024, MME-CoT, zhu2026mmrv, li2026mmrlife, wang2026thinkwatchingonlinestreaming, jin2025omnireward}, enabling reasoning over text and visual modalities. Recent study highlights the significant differences in the improvement of CoT across tasks and reveal limitations of current CoT paradigms~\cite{tocot}. 
However, there is still a lack of systematic analysis of multimodal CoT.

\noindent \textbf{Multimodal Reasoning}.
Reasoning models such as OpenAI's o1~\cite{openai2024o1}, DeepSeek R1~\cite{deepseekr1}, and QwQ~\cite{qwq32b} achieve strong results on text reasoning. Building on this, models like LLaVA-o1~\cite{xu2024llava}, R1-Onevision~\cite{R1-Onevision}, MM-Eureka~\cite{meng2025mm}, OpenVLThinker~\cite{OpenVLThinker}, VL-Rethinker~\cite{wang2025vl}, VLM-R1~\cite{shen2025vlm}, and X-Reasoner~\cite{liu2025xreasonergeneralizablereasoningmodalities} extend reasoning to multimodal tasks, showing improvements in mathematical reasoning and long CoT capabilities. However, most of them lack validation across broader multimodal tasks.

\noindent \textbf{Thinking with Image}. Integrating visual modality into CoT reasoning process enables `\textit{thinking with images}' that transcends purely textual reasoning \cite{imagine_while_reasoning,grit,pixel_reasoner,reasoning_in_the_dark,reinforcing_spatial_reasoning,mitigate_model_imbalance,li2026aee}. Models can be empowered through explicit tool-use for visual manipulations, such as cropping and zooming \cite{twi_DeepEyes,twi_VGR}.  Additionally, code-based operation provides even greater flexibility and versatility for diverse visual reasoning scenarios \cite{twi_PyVision,twi_Visual_ARFT, jin2026pixels}.

\section{Conclusion}

In this paper, we present a comprehensive study on the strengths and limitations of multimodal CoT reasoning.
Our findings reveal that: (1) CoT's efficacy is task-dependent and requires selective application; (2) current open-source models show marginal gains, likely due to an overemphasis on mathematical reasoning; and (3) visual reasoning remains a bottleneck, characterized by a ``\textit{Look Light, Think Heavy}'' pattern where visual reflection diminishes compared to verbal reflection. 
To address these limitations, a promising path forward is reasoning with visual reflection and external tools.

\section*{Limitations}
\label{Limitations}

Despite our comprehensive analysis of multimodal CoT reasoning, our study faces two limitations.
First, due to computational constraints, we evaluate only a subset of widely adopted datasets (1–3 per task) across 12 multimodal tasks, and conduct experiments on 14 general models and 8 reasoning models.
While this setup covers a wide range of capabilities, it may not fully capture the diversity of multimodal tasks. 
In future work, we plan to expand our evaluation by including more datasets, testing a wider variety of models, and extending our analysis to video-related perception and reasoning tasks.
Second, although our findings uncover a fundamental limitation of current multimodal CoT, namely the “\textit{Look Light, Think Heavy}” phenomenon.
Inspired by o3, we attempt to prompt GPT-4.1 to perform multimodal tool-enhanced CoT reasoning.
However, we find that even a strong model like GPT-4.1 tends to favor text-oriented tools \cite{, hao-etal-2025-evaluating}, such as numerical calculators, rather than leveraging visual tools that could enhance image understanding and reasoning, revealing a lack of visual tool-use awareness in current models.
This highlights the need for future MLLMs to more effectively integrate visual tools into the CoT reasoning process.

We also propose two promising directions to address this limitation:
(1) \textbf{Reasoning with Visual Reflections}: As shown in Figure \ref{fig:case1}, when presented with images where key information is obscured by mosaics, o3 is able to first recognize the visual ambiguity, then zoom in on the occluded region, analyze the lack of detail, and ultimately conclude that the visual input is insufficient, resulting in an appropriate refusal to answer.
Explicitly cropping and zooming in on and revisiting critical visual areas facilitates deeper visual reflection.
(2) \textbf{Reasoning with External Tools}: As shown in Figure \ref{fig:case2}, when confronted with complex visual inputs such as the Eight Queens puzzle, the model first invokes an external visual tool to accurately identify the positions of the chess pieces, and then executes algorithmic code to complete the task.
Reasoning with external tools significantly expands the capability boundaries of MLLMs.

\section*{Acknowledgements}
This work was supported by the National Natural Science Foundation of China (No.U24A20335, No.62406321), Beijing Natural Science Foundation (L243006), and the independent research project of the Key Laboratory of Cognition and Decision Intelligence for Complex Systems.

\bibliography{custom}

\clearpage
\appendix

\section{Task Details}
\label{Task Details}

\begin{table*}[t]
\centering
\caption{Overview of the evaluation benchmark. We categorize the 12 tasks into Perception and Reasoning, listing the source datasets and the corresponding sample sizes (in parentheses) for each.}
\label{tab:task_overview}
\resizebox{0.8\linewidth}{!}{
\begin{tabular}{lll}
\toprule
\textbf{Category} & \textbf{Task} & \textbf{Dataset(s)} \\
\midrule
Perception & Comprehensive Evaluation & MME (200), MMStar (200), MMT-Bench (200) \\
Perception & OCR & TextVQA (200), OCRBench (200) \\
Perception & Visual Grounding & RefCOCO (150), RefCOCOg (150) \\
Perception & Hallucination Detection & HallucinationBench (250), POPE (250) \\
Perception & Knowledge-Based VQA & A-OKVQA (200) \\
Perception & Object Counting & Super-CLEVR (200) \\
\midrule
Reasoning & Mathematical Reasoning & MathVerse (200) \\
Reasoning & Scientific Reasoning & MMMU (200) \\
Reasoning & Logical Reasoning & PuzzleVQA (200) \\
Reasoning & Algorithmic Reasoning & AlgoPuzzleVQA (200) \\
Reasoning & Spatial Reasoning & SpatialEval (200) \\
Reasoning & Multi-Image Reasoning & MUIRBench (200) \\
\bottomrule
\end{tabular}
}
\end{table*}

\subsection{Multimodal Perception Tasks}

Table \ref{tab:task_overview} provides an overview of the datasets and sample sizes used for the tasks we evaluated. More detailed information is provided below.

\begin{figure*}[h]
    \centering
    \includegraphics[width=0.80\linewidth]{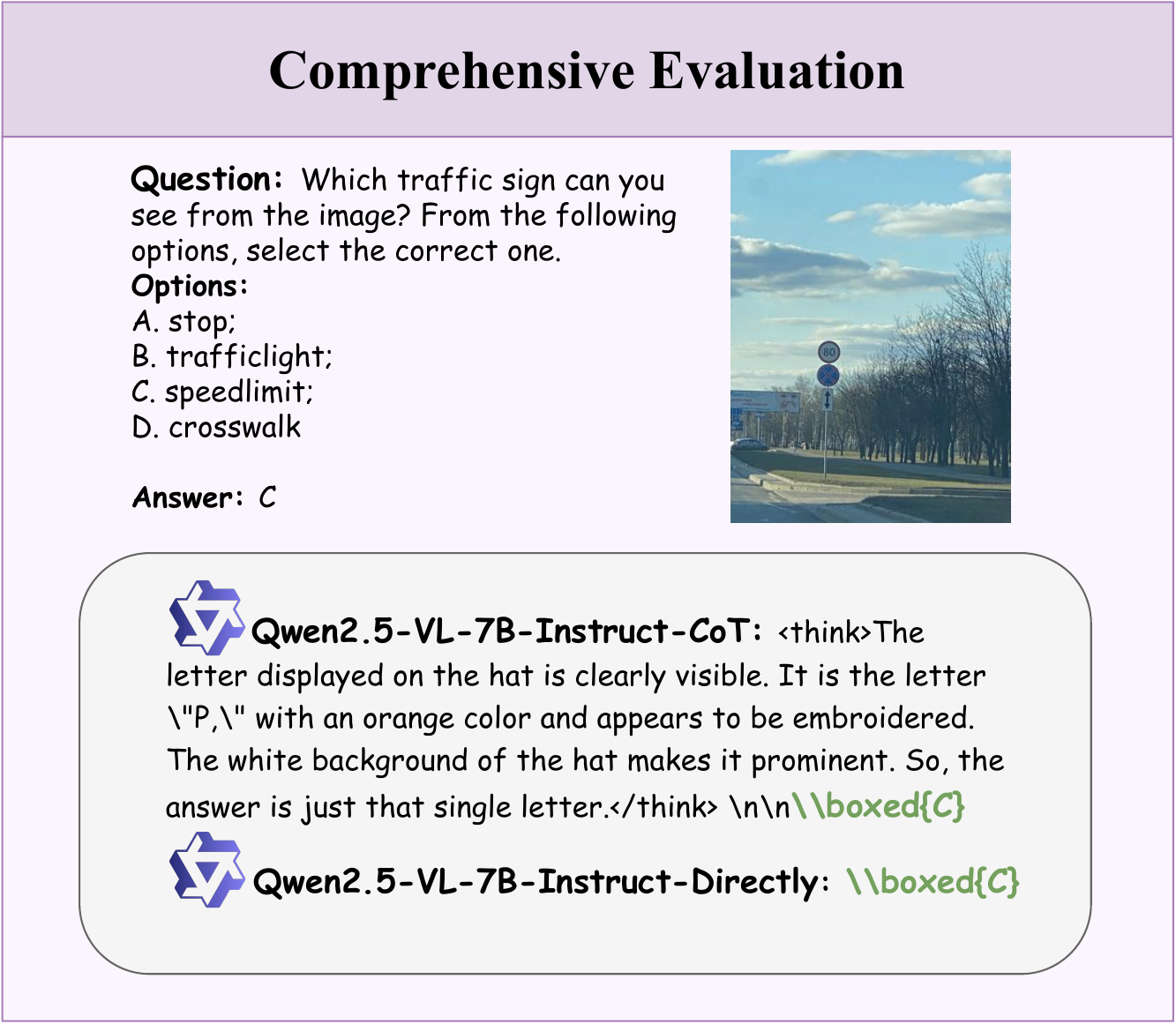}
    \caption{An example of the comprehensive evaluation task with both direct and CoT responses.}
    \label{fig:comprehensive_case}
\end{figure*}

\textbf{Comprehensive Evaluation}. Comprehensive evaluation \cite{DBLP:journals/corr/abs-2411-15296} refers to the systematic assessment of MLLMs across a broad range of capabilities. 
We select three benchmarks (MME~\cite{fu2023mme}, MMStar~\cite{chen2024we}, and MMT-Bench~\cite{ying2024mmt}) and sample 200 questions from each to construct the evaluation set. MME provides a broad assessment of model performance in multitask settings. MMStar addresses issues related to visual independence and data leakage. MMT-Bench focuses on real-world applications and everyday visual content.
Figure~\ref{fig:comprehensive_case} presents an example comparing the direct response and the CoT response generated by Qwen2.5-VL-7B-Instruct.

\begin{figure*}[h]
    \centering
    \includegraphics[width=0.80\linewidth]{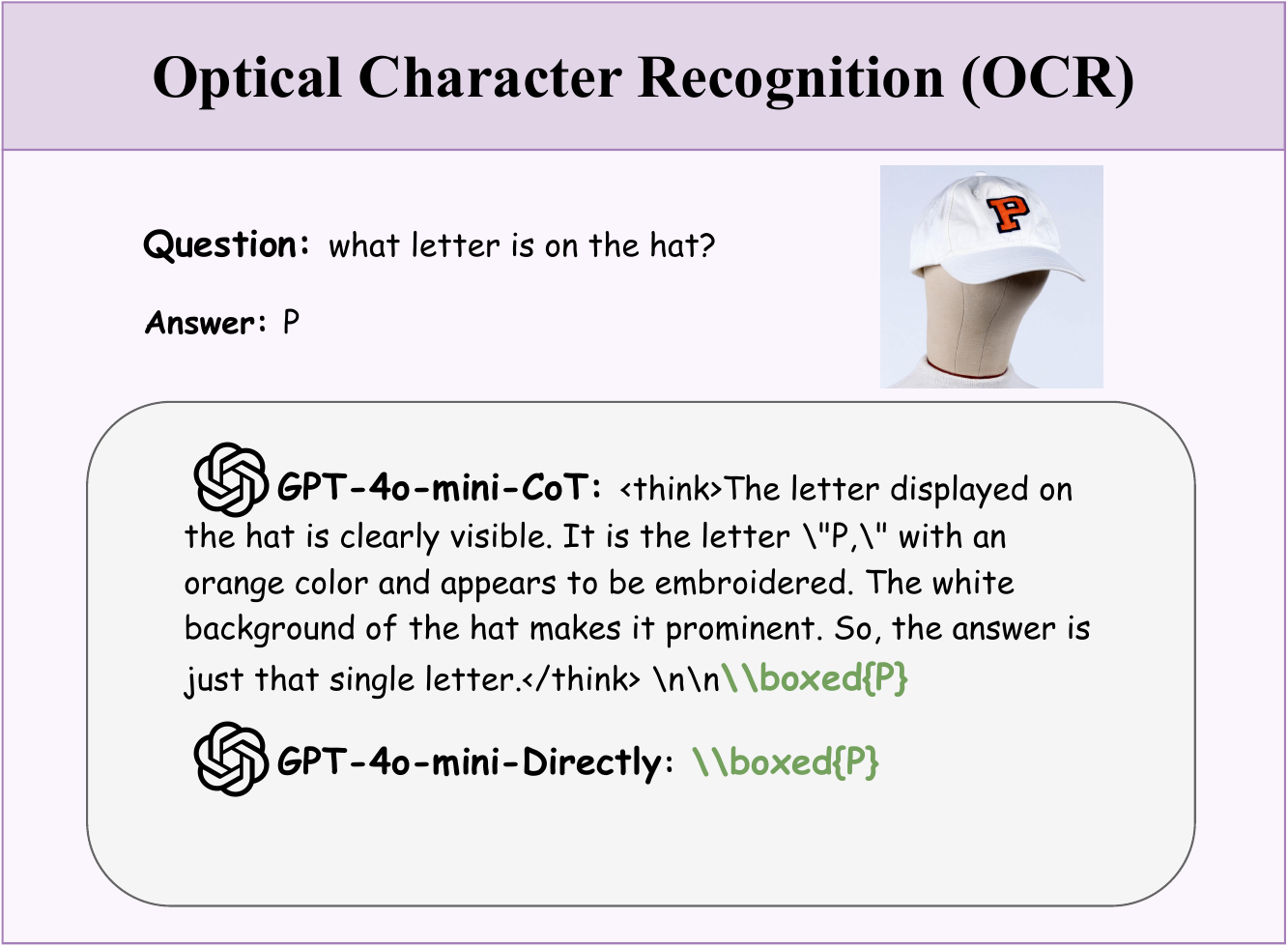}
    \caption{An example of the OCR task with both direct and CoT responses.}
    \label{fig:ocr_case}
\end{figure*}

\textbf{Optical Character Recognition (OCR)}. OCR is the task of automatically detecting and transcribing textual content from images, evaluating fine-grained visual perception and the accuracy of cross-modal transcription. We select TextVQA~\cite{singh2019towards} and OCRBench~\cite{liu2024ocrbench} to construct the evaluation set. TextVQA focuses on visual question answering that requires understanding text in real-world photographs. OCRBench expands the scope to various data types such as charts, maps, and web pages.
Figure~\ref{fig:ocr_case} presents an example comparing the direct response and the CoT response generated by GPT-4o-mini.

\begin{figure*}[h]
    \centering
    \includegraphics[width=0.80\linewidth]{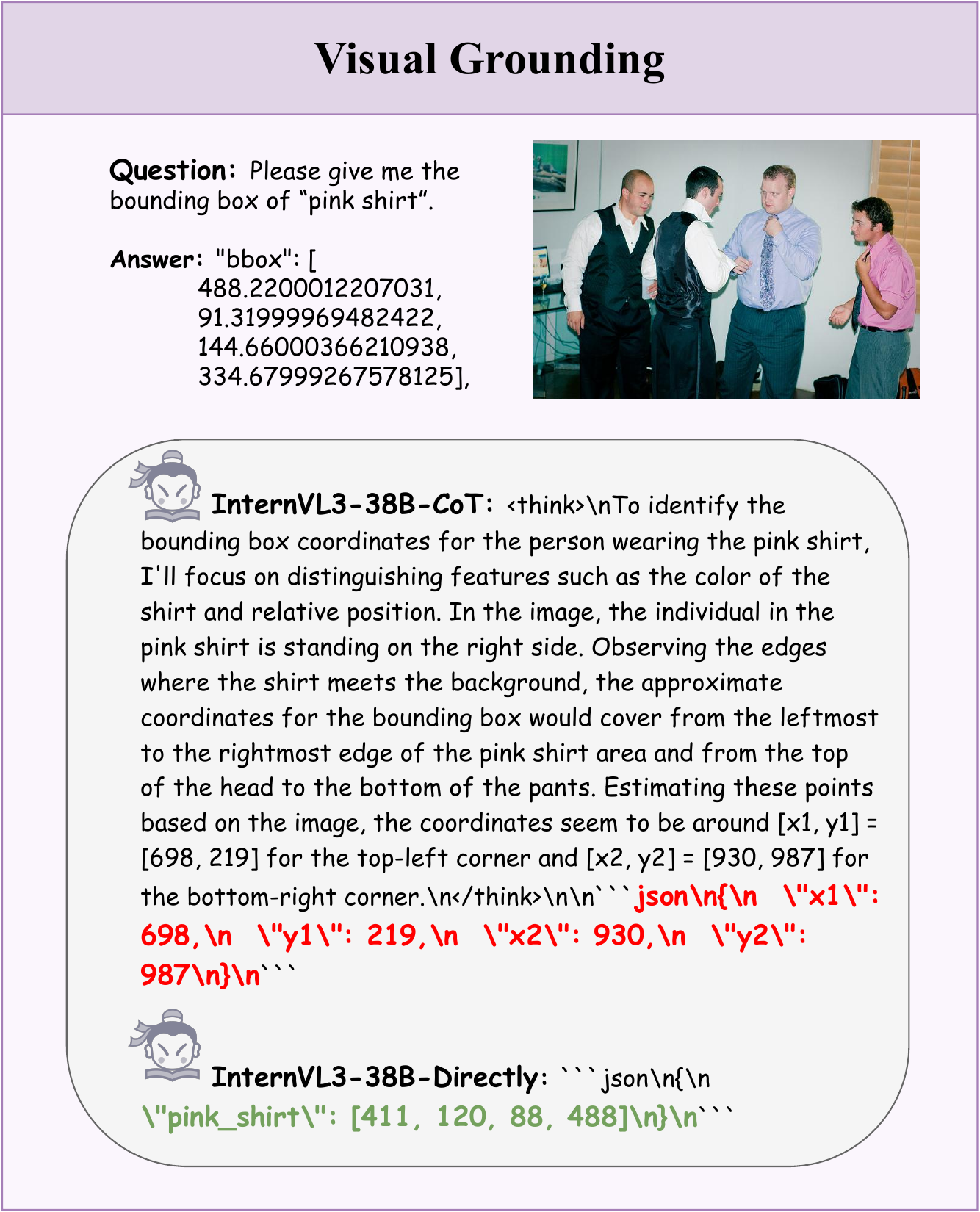}
    \caption{An example of the visual grounding task with both direct and CoT responses.}
    \label{fig:visual grounding_case}
\end{figure*}

\textbf{Visual Grounding}.
Visual grounding involves localizing the referent of a textual description (\textit{e.g.}, \textit{``man in back''}) in an image by predicting a corresponding bounding box. 
It aims to evaluate the ability of models to align cross-modal information and to accurately recognize and localize visual entities.
We sample 150 instances each from the widely used RefCOCO and RefCOCOg~\cite{kazemzadeh2014referitgame} benchmarks to construct the task set.
Figure~\ref{fig:visual grounding_case} presents an example comparing the direct response and the CoT response generated by InternVL3-38B.

\begin{figure*}[h]
    \centering
    \includegraphics[width=0.80\linewidth]{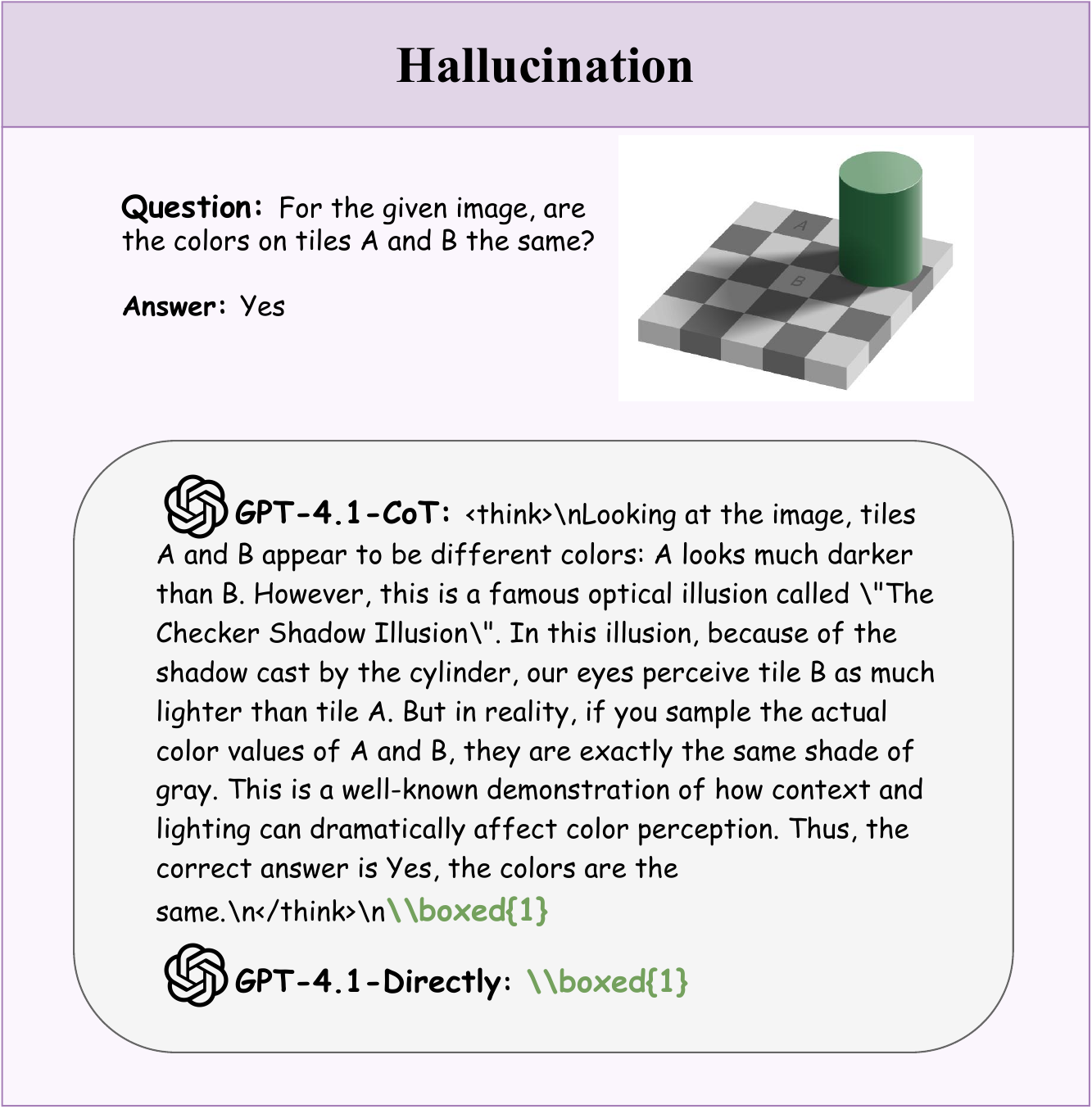}
    \caption{An example of the hallucination task with both direct and CoT responses.}
    \label{fig:hallucination_case}
\end{figure*}

\textbf{Hallucination}. Multimodal hallucination evaluation focuses on assessing the phenomenon of models to generate content that is not grounded in the input modalities (especially the visual modality), thereby measuring the factual consistency between generated outputs and the given multimodal evidence. We sample 250 tasks each from HallusionBench~\cite{liu2023hallusionbench} and POPE~\cite{li2023evaluating} to construct the evaluation set.
Figure~\ref{fig:hallucination_case} presents an example comparing the direct response and the CoT response generated by GPT-4.1.

\begin{figure*}[h]
    \centering
    \includegraphics[width=0.80\linewidth]{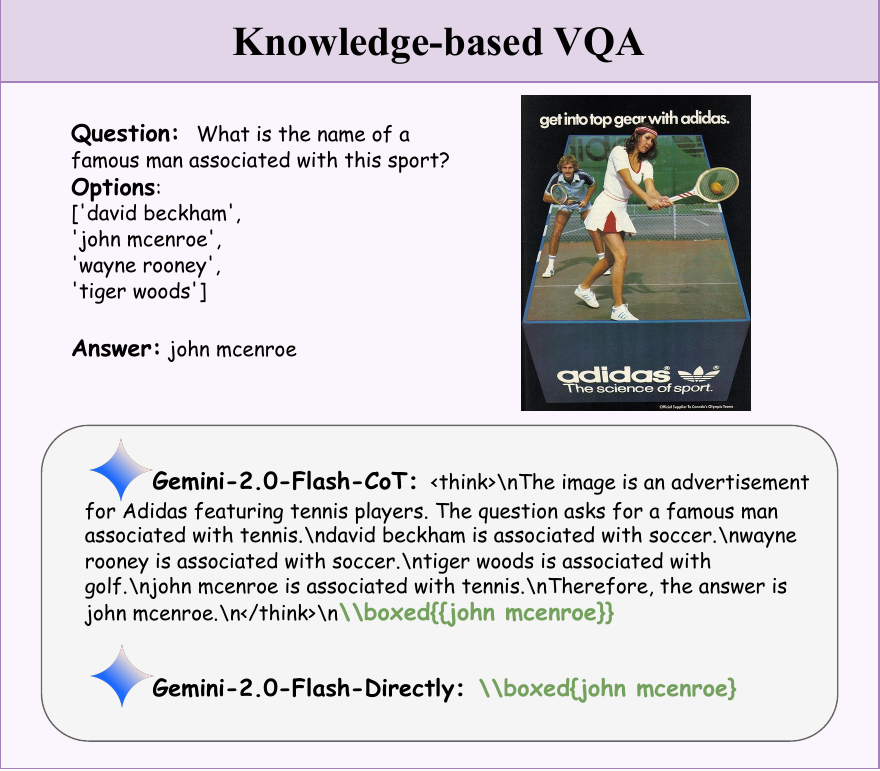}
    \caption{An example of the knowledge-base VQA task with both direct and CoT responses.}
    \label{fig:knowledge_case}
\end{figure*}

\textbf{Knowledge-based VQA}. Unlike standard VQA tasks, knowledge-based VQA is designed to assess a model’s ability to answer questions that require commonsense and world knowledge beyond what is directly observable in the image. 
We sample 200 questions from A-OKVQA~\cite{schwenk2022okvqa} as the test set.
Figure~\ref{fig:knowledge_case} presents an example comparing the direct response and the CoT response generated by Gemini-2.0-Flash.

\begin{figure*}[h]
    \centering
    \includegraphics[width=0.80\linewidth]{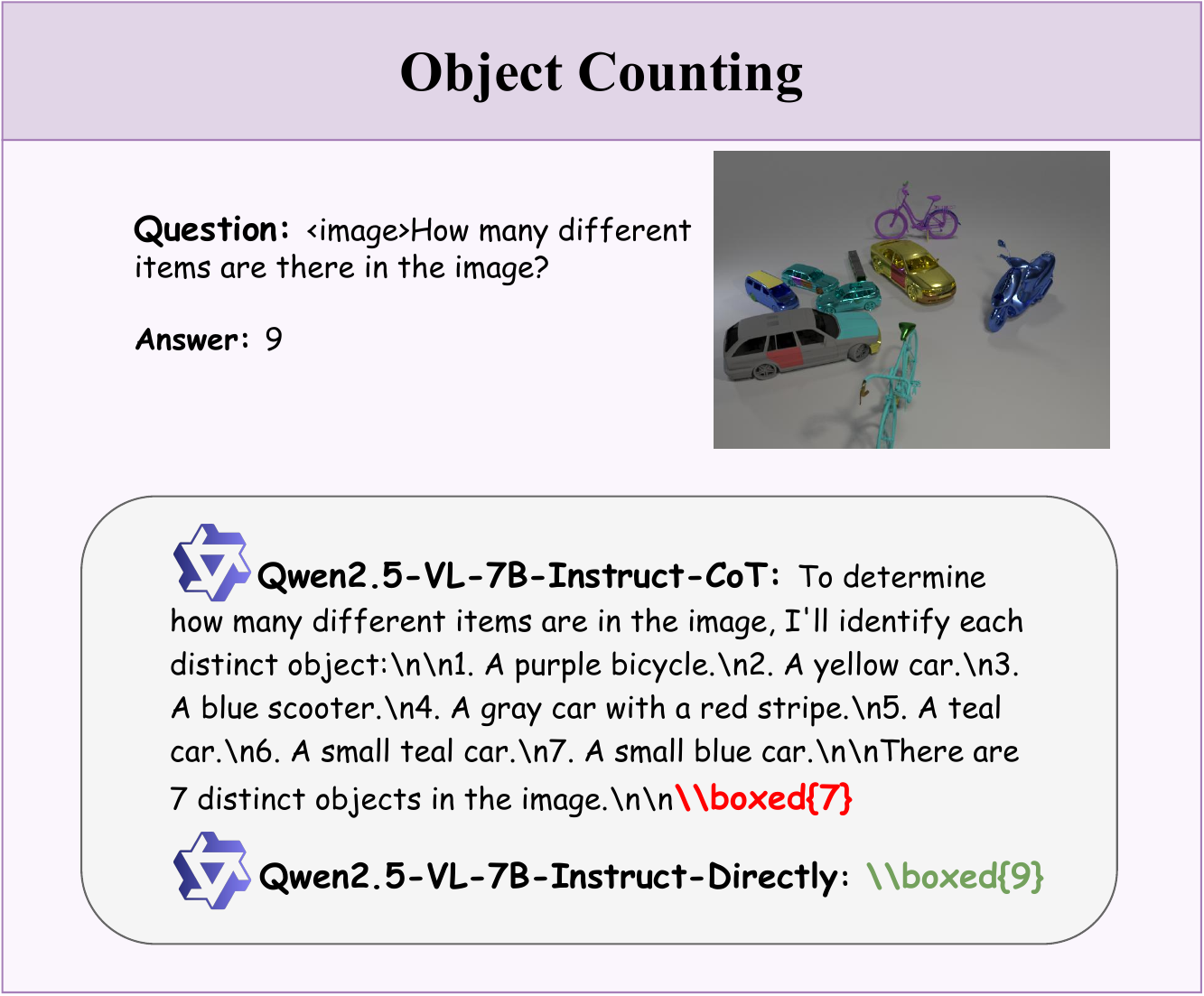}
    \caption{An example of the object counting task with both direct and CoT responses.}
    \label{fig:counting_case}
\end{figure*}

\textbf{Object Counting}. This task requires the model to perceive the number of distinct entities in an image (\textit{e.g.},``\textit{How many different items are there in the image?}''), assessing the accuracy of visual understanding and perception. We select 200 samples from Super-CLEVR~\cite{li2023super} as the test set. 
This dataset is an enhanced version of the classic counting benchmark CLEVR~\cite{johnson2017clevr}, extending object types from simple geometric shapes to more realistic entities such as bicycles.
Figure~\ref{fig:counting_case} presents an example comparing the direct response and the CoT response generated by Qwen2.5-VL-7B-Instruct.

\subsection{Multimodal Reasoning Tasks}

\begin{figure*}[h]
    \centering
    \includegraphics[width=0.80\linewidth]{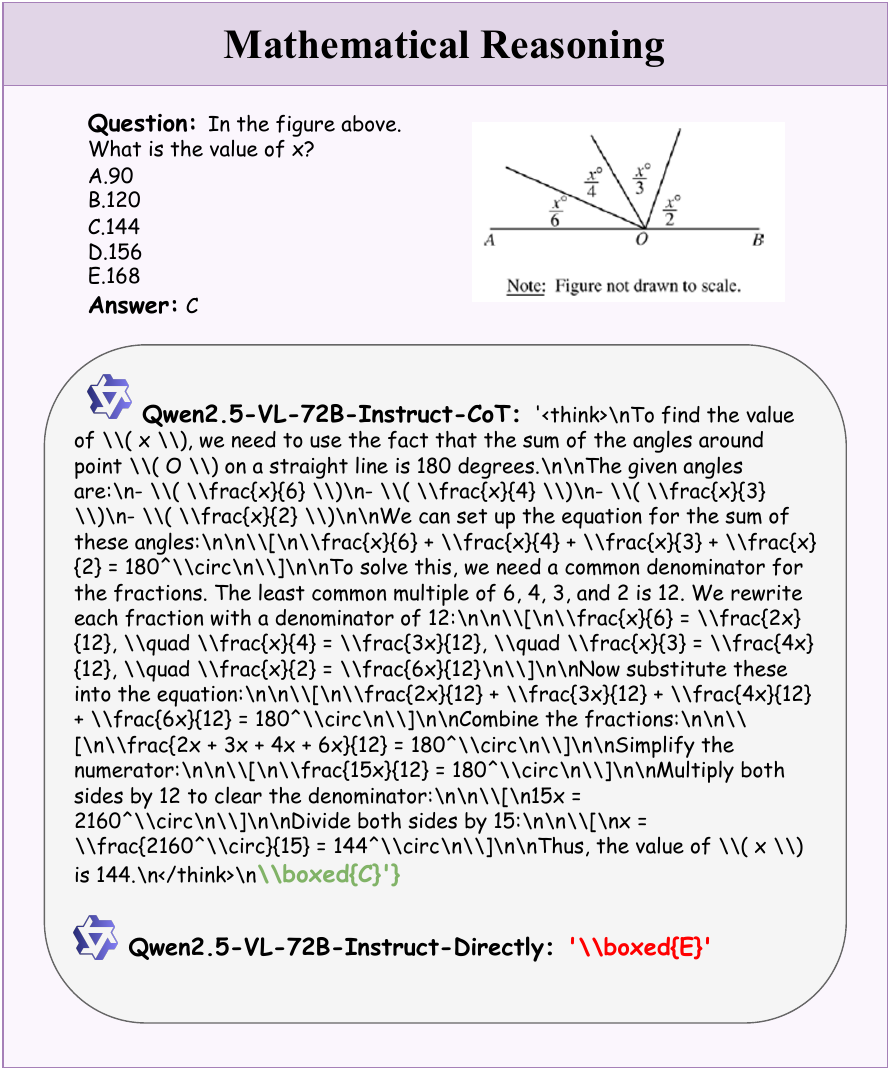}
    \caption{An example of the mathematical reasoning task with both direct and CoT responses.}
    \label{fig:math_case}
\end{figure*}

\textbf{Mathematical Reasoning}. Mathematical reasoning tasks assess a model's ability to understand and solve problems involving mathematical concepts, multi-step inference, and precise computation. It is one of the most actively studied areas in multimodal reasoning. We sample 200 tasks from the MathVerse~\cite{MathVerse} benchmark to construct the test set. 
Figure~\ref{fig:math_case} presents an example comparing the direct response and the CoT response generated by Qwen2.5-VL-72B-Instruct.

\begin{figure*}[h]
    \centering
    \includegraphics[width=0.80\linewidth]{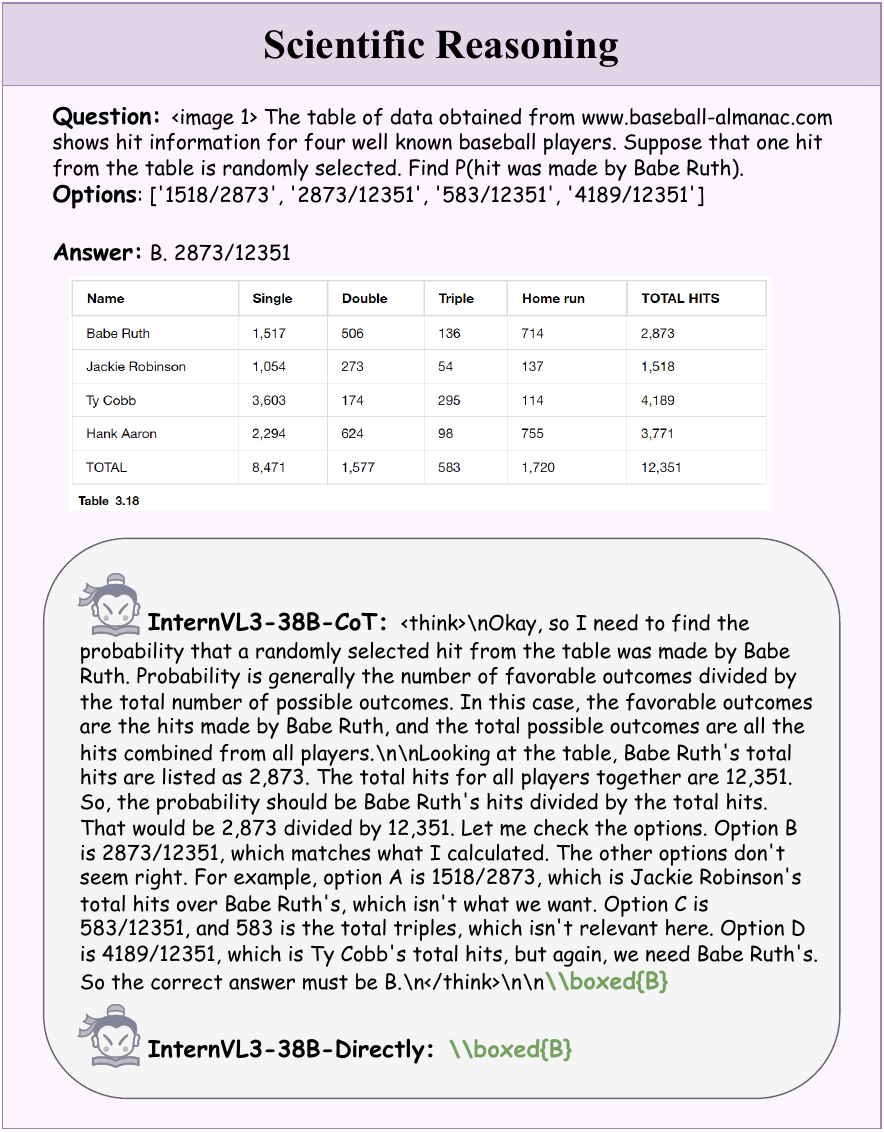}
    \caption{An example of the scientific reasoning task with both direct and CoT responses.}
    \label{fig:scientific_case}
\end{figure*}

\textbf{Scientific Reasoning}. These tasks evaluate the ability of models to comprehend and reasoning for information from multiple modalities (\textit{e.g.}, text, charts, and images) to answer questions that require scientific knowledge. We sample 200 tasks from MMMU~\cite{mmmu} for evaluation, which covers graduate-level, multimodal science questions across diverse disciplines.
Figure~\ref{fig:scientific_case} presents an example comparing the direct response and the CoT response generated by InternVL3-38B.

\begin{figure*}[h]
    \centering
    \includegraphics[width=0.80\linewidth]{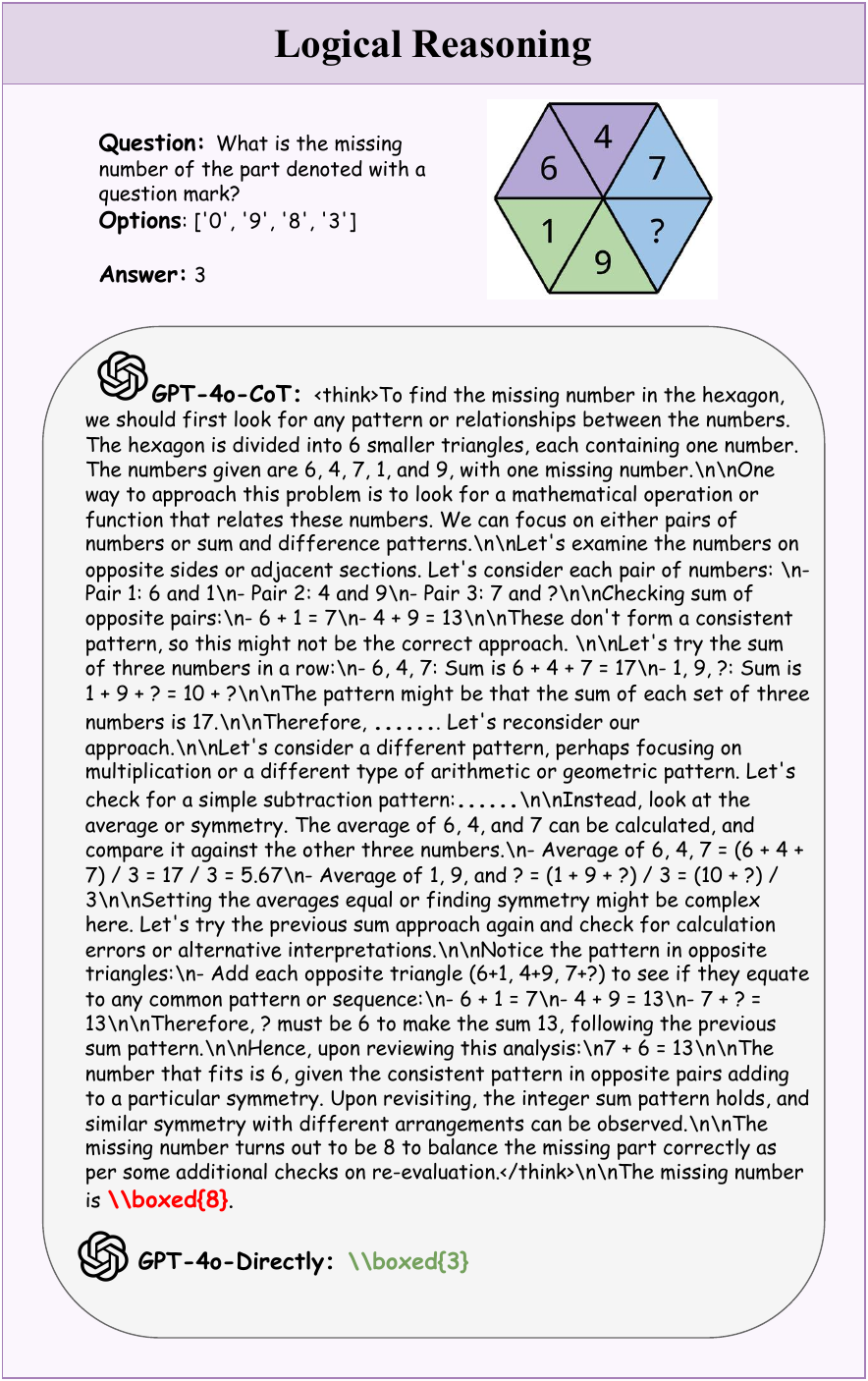}
    \caption{An example of the logical reasoning task with both direct and CoT responses.}
    \label{fig:logical_case}
\end{figure*}

\textbf{Logical Reasoning}. These tasks evaluate the capacity to reason and apply logical principles across multiple modalities, requiring models to draw conclusions, make predictions, recognize patterns, and solve problems or puzzles based on multimodal inputs and given premises. We select PuzzleVQA~\cite{chia2024puzzlevqa}, a visual puzzle benchmark, and sample 200 tasks to construct the evaluation set.
Figure~\ref{fig:logical_case} presents an example comparing the direct response and the CoT response generated by GPT-4o.

\begin{figure*}[h]
    \centering
    \includegraphics[width=0.80\linewidth]{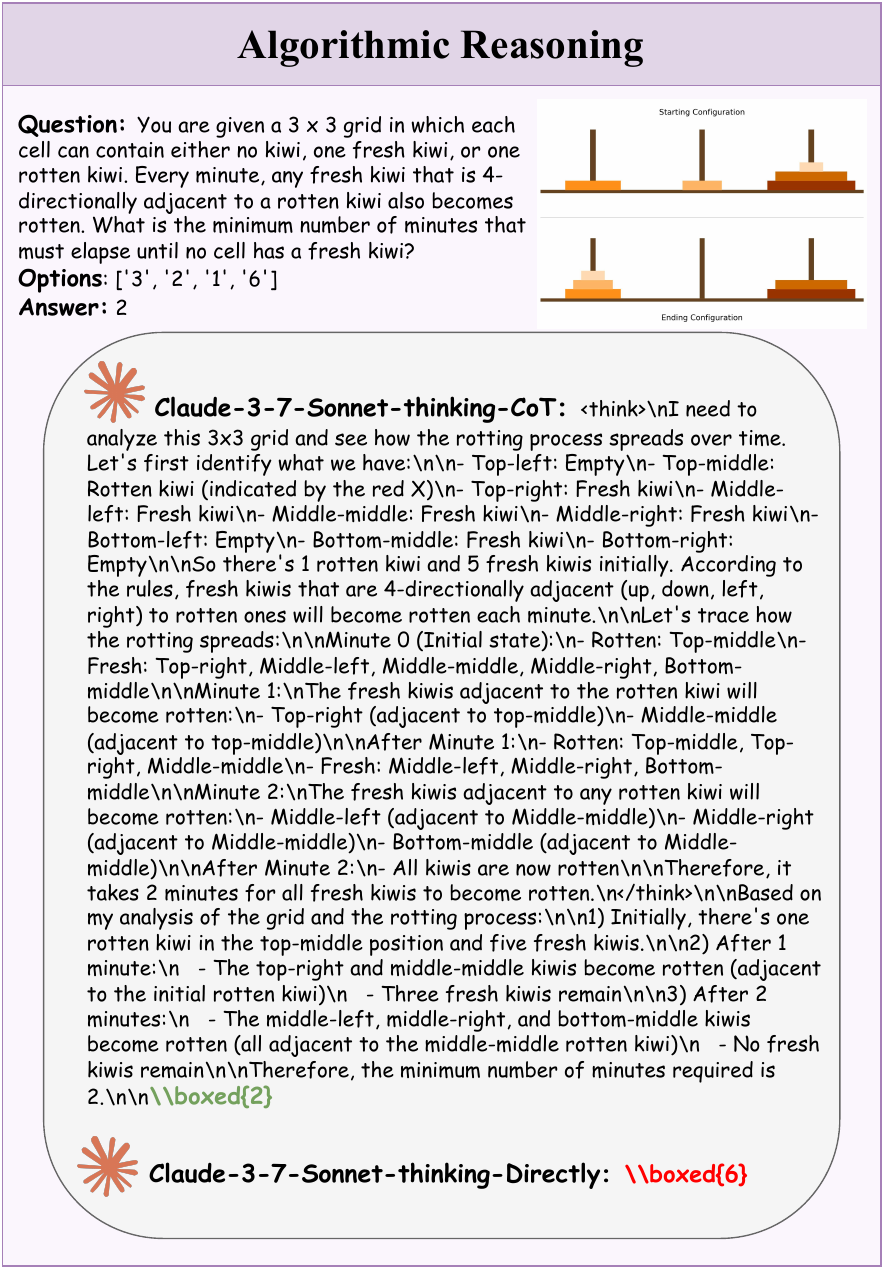}
    \caption{An example of the algorithmic reasoning task with both direct and CoT responses.}
    \label{fig:algorithmic_case}
\end{figure*}

\textbf{Algorithmic Reasoning}. Algorithmic reasoning tasks assess a model’s ability to understand and reason through step-by-step computational procedures in a multimodal setting. These tasks cover areas such as graph theory, combinatorics, and search problems (\textit{e.g.}, the eight queens problem). We select 200 tasks from the algorithmic dataset AlgoPuzzleVQA~\cite{ghosal2024language} to construct the evaluation set.
Figure~\ref{fig:algorithmic_case} presents an example comparing the direct response and the CoT response generated by Claude-3-7-Sonnet-Thinking.

\begin{figure*}[h]
    \centering
    \includegraphics[width=0.80\linewidth]{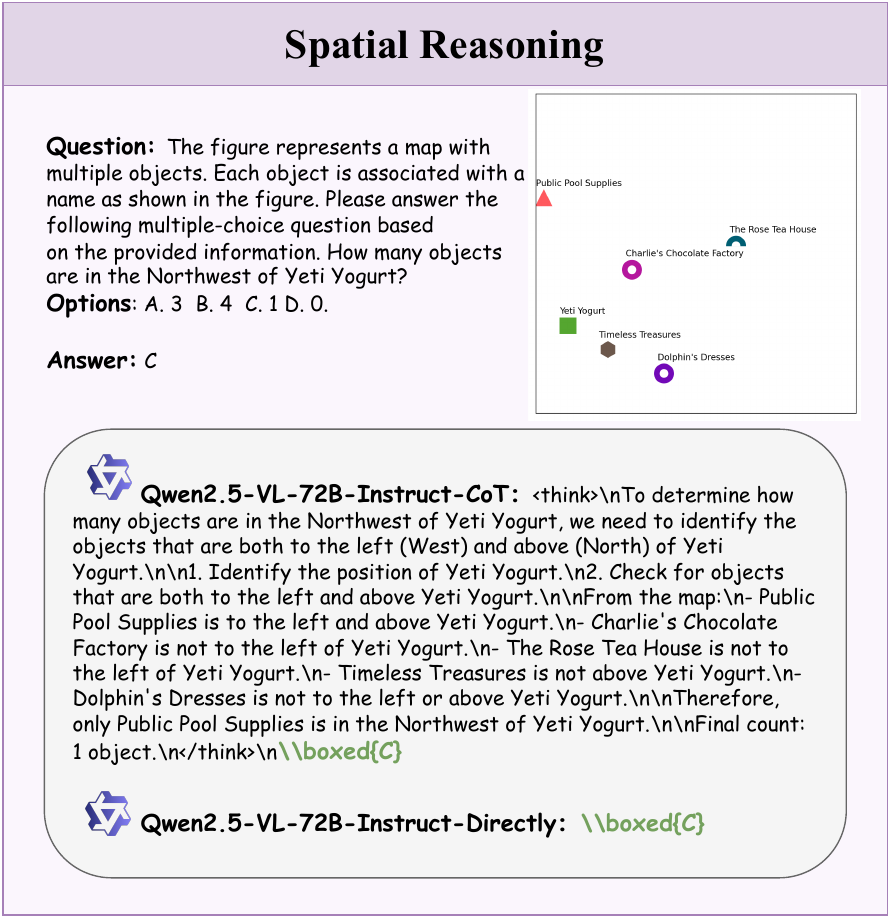}
    \caption{An example of the spatial reasoning task with both direct and CoT responses.}
    \label{fig:spatial_case}
\end{figure*}

\textbf{Spatial Reasoning}. Spatial reasoning tasks assess the ability to understand and analyze spatial relationships between objects, including position, orientation, distance, and movement, often requiring inference from visual inputs to solve problems related to navigation, assembly, or geometric reasoning. We sample 200 tasks from SpatialEval~\cite{wang2024picture} for evaluation.
Figure~\ref{fig:spatial_case} presents an example comparing the direct response and the CoT response generated by Qwen2.5-VL-72B-Instruct.

\begin{figure*}[h]
    \centering
    \includegraphics[width=0.80\linewidth]{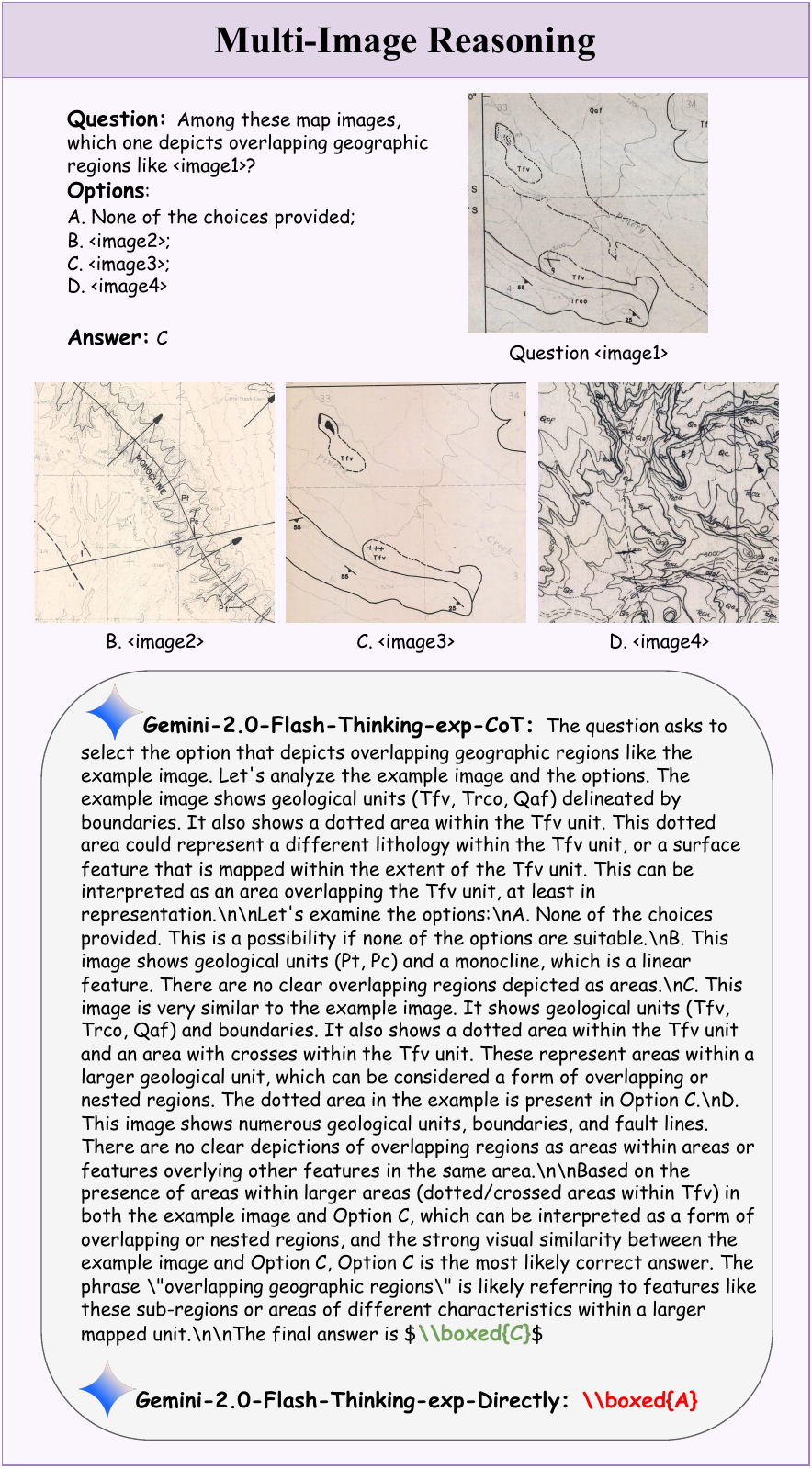}
    \caption{An example of the multi-image reasoning task with both direct and CoT responses.}
    \label{fig:multi-image_case}
\end{figure*}

\textbf{Multi-Image Reasoning}. 
These tasks evaluate the ability to jointly analyze information from multiple images to perform complex reasoning for the problem, such as comparison, temporal or causal inference, and cross-image consistency reasoning, often requiring a holistic understanding that goes beyond single-image perception. We sample 200 tasks in MUIRBENCH~\cite{wang2024muirbench} for evaluation.
Figure~\ref{fig:multi-image_case} presents an example comparing the direct response and the CoT response generated by Gemini-2.0-Flash-Thinking.

\vspace*{\fill}
\vspace*{\fill}
\clearpage

\clearpage

\clearpage

\clearpage

\clearpage

\clearpage

\clearpage

\clearpage

\clearpage

\clearpage

\clearpage

\clearpage

\section{Evaluation Details}
\label{Prompts}
For \textbf{non-reasoning models}, we evaluate 14 models: Qwen2.5-VL (7B/32B/72B-Instruct)~\cite{qwen25vl}, Qwen2-VL-72B-Instruct~\cite{wang2024qwen2}, Intern3-VL (8B/14B/38B)~\cite{zhu2025internvl3}, Gemma-3 (4B/12B)~\cite{gemma3}, Mistral-Small-3.1-24B-Instruct~\cite{mistral-small}, GPT-4o-mini, GPT-4o~\cite{openai2024gpt4o}, GPT-4.1~\cite{openai2025gpt41} and Gemini-2.0-Flash~\cite{team2023gemini}.
For \textbf{reasoning models}, we evaluate 8 models: VL-Rethinker (7B/72B)~\cite{wang2025vl} and MM-Eureka (7B/32B)~\cite{meng2025mm} (Qwen2.5-VL Based), QVQ-72B-Preview~\cite{QVQ} (Qwen2-VL Based), Skywork-R1V2-38B~\cite{SkyworkR1V2}, Kimi-VL-A3B-Thinking~\cite{kimiteam2025kimivltechnicalreport}, and Gemini-2.0-Flash-Thinking~\cite{team2023gemini} (Gemini-2.0-Flash Based).
Considering the large number of experiments and limited computational resources, we consistently adopt the performance@1 setting.
We use all models in compliance with their respective licenses.
The prompt varies according to the task type. For models with specific prompts, we retain their original prompt design; otherwise, a standardized prompt is adopted.

\begin{table*}[h]
\caption{    
Prompt for the comprehensive evaluation task.
}
\label{appendix:prompt_comprehensive}
\centering
\resizebox{0.99\linewidth}{!}{    \small
    \begin{tabular}{p{\linewidth}}
        \toprule
        \underline{\textbf{Prompt for Comprehensive Evaluation Task}} \\
        \vspace{-2mm}
        \hl{\textbf{\textsc{Direct Answer}:}} \\ 
Please generate the answer directly, and it MUST be enclosed in \textbackslash boxed\{\}.

YN Prompt:

Based on the image, answer the following question in [[OUTPUT FORMAT]]: \colorbox{cyan!20}{\{question\}}

[[OUTPUT FORMAT]]

Format your answer as follows:

If the answer is Yes, directly give the final answer in the following format: \textbackslash boxed\{\{Y\}\}.

If the answer is Yes, directly give the final answer in the following format: \textbackslash boxed\{\{Y\}\}.

[[END OF OUTPUT FORMAT]]

MC prompt:

Based on the image, select the correct option of the following question in [[OUTPUT FORMAT]]: \colorbox{cyan!20}{\{question\}}

[[OUTPUT FORMAT]]

Format your answer as follows:

If the correct option letter is X, directly give the final correct letter in the following format: \textbackslash boxed\{\{X\}\}.

[[END OF OUTPUT FORMAT]]

\hl{\textbf{\textsc{CoT}:}} \\ 

You FIRST think about the reasoning process as an internal monologue and then provide the final answer. The reasoning process MUST BE enclosed within <think> </think> tags. The final answer MUST BE put in \textbackslash boxed\{\}.

YN Prompt:

Based on the image, answer the following question in [[OUTPUT FORMAT]]: \colorbox{cyan!20}{\{question\}}

Let's think step by step.

[[OUTPUT FORMAT]]

Format your answer as follows:

Your thinking process enclosed within <think> </think> tags.

If the answer is Yes, give the final answer in the following format: \textbackslash boxed\{\{Y\}\}.

If the answer is No, give the final answer in the following format: \textbackslash boxed\{\{Y\}\}.

[[END OF OUTPUT FORMAT]]

MC prompt:

Based on the image, select the correct option of the following question in [[OUTPUT FORMAT]]: \colorbox{cyan!20}{\{question\}}

Let's think step by step.

[[OUTPUT FORMAT]]

Format your answer as follows:

Your thinking process enclosed within <think> </think> tags.

If the correct option letter is X, give the final correct letter in the following format: \textbackslash boxed\{\{X\}\}.

[[END OF OUTPUT FORMAT]] \\
         \bottomrule
\end{tabular}}
    
\end{table*}

\begin{table*}[h]
\caption{    
Prompt for the OCR task.
}
\label{appendix:prompt_ocr}
\centering
\resizebox{0.99\linewidth}{!}{    \small
    \begin{tabular}{p{\linewidth}}
        \toprule
        \underline{\textbf{Prompt for OCR Task}} \\
        \vspace{-2mm}
        \hl{\textbf{\textsc{Direct Answer}:}} \\ Please generate the answer directly, and it MUST be enclosed in \textbackslash boxed\{\}. \\

Please try to answer the question with short words or phrases if possible.

Question: \colorbox{cyan!20}{\{question\}}   \\

\hl{\textbf{\textsc{CoT}:}} \\ You FIRST think about the reasoning process as an internal monologue and then provide the final answer. The reasoning process MUST BE enclosed within <think> </think> tags. The final answer MUST BE put in \textbackslash boxed\{\}. \\

Please try to answer the question with short words or phrases if possible.

Question: \colorbox{cyan!20}{\{question\}}   \\

        \bottomrule
    \end{tabular}}
    
\end{table*}

\begin{table*}[h]
\caption{    
Prompt for the visual grounding task.
}
\label{appendix:prompt_visual_grounding}
\centering
\resizebox{0.99\linewidth}{!}{    \small
    \begin{tabular}{p{\linewidth}}
        \toprule
        \underline{\textbf{Prompt for Visual Grounding Task}} \\
        \vspace{-2mm}
        \hl{\textbf{\textsc{Direct Answer}:}} \\ Please answer the option's letter from the given choices directly, and it MUST be enclosed in \textbackslash boxed\{\}. \\

Please provide the bounding box coordinate of the region this sentence describes. \\
Question: \colorbox{cyan!20}{\{question\}} \\

Format your answer as follows: \\
output its bbox coordinates using JSON format. \\

\hl{\textbf{\textsc{CoT}:}} \\ You FIRST think about the reasoning process as an internal monologue and then provide the final answer with the option's letter from the given choices directly. The reasoning process MUST BE enclosed within <think> </think> tags. The final answer MUST BE put in \textbackslash boxed\{\}. \\

Please provide the bounding box coordinate of the region this sentence describes. \\
Question: \colorbox{cyan!20}{\{question\}} \\
Let's think step by step. \\
Format your answer as follows: \\
output its bbox coordinates using JSON format. \\

        \bottomrule
    \end{tabular}}
    
\end{table*}

\begin{table*}[h]
\caption{    
Prompt for the hallucination task.
}
\label{appendix:prompt_hallucination}
\centering
\resizebox{0.99\linewidth}{!}{    \small
    \begin{tabular}{p{\linewidth}}
        \toprule
        \underline{\textbf{Prompt for Hallucination Task}} \\
        \vspace{-2mm}
        \hl{\textbf{\textsc{Direct Answer}:}} \\ Please generate the answer directly, and it MUST be enclosed in \textbackslash boxed\{\}. \\

Answer the following question. \\
Question: \colorbox{cyan!20}{\{question\}}   \\
The answer is Yes or No. \\

Format your answer as follows: \\
If the answer is Yes, directly give the final answer in the following format: \textbackslash boxed\{1\}. \\
If the answer is No, directly give the final answer in the following format: \textbackslash boxed\{0\}. \\

\hl{\textbf{\textsc{CoT}:}} \\ You FIRST think about the reasoning process as an internal monologue and then provide the final answer. The reasoning process MUST BE enclosed within <think> </think> tags. The final answer MUST BE put in \textbackslash boxed\{\}. \\

Answer the following question. \\
Question: \colorbox{cyan!20}{\{question\}}   \\
The answer is Yes or No.\\
Let's think step by step.\\

Format your answer as follows:\\
Your thinking process enclosed within <think> </think> tags.\\
If the answer is Yes, give the final answer in the following format: \textbackslash boxed\{1\}.\\
If the answer is No, give the final answer in the following format: \textbackslash boxed\{0\}.\\

        \bottomrule
    \end{tabular}}
    
\end{table*}

\begin{table*}[h]
\caption{    
Prompt for the knowledge-based VQA task.
}
\label{appendix:prompt_knowledeg}
\centering
\resizebox{0.99\linewidth}{!}{    \small
    \begin{tabular}{p{\linewidth}}
        \toprule
        \underline{\textbf{Prompt for Knowledge-Based VQA Task}} \\
        \vspace{-2mm}
        \hl{\textbf{\textsc{Direct Answer}:}} \\ Please generate the answer directly, and it MUST be enclosed in \textbackslash boxed\{\}. \\
Question: \colorbox{cyan!20}{\{question\}}   \\ 
Options: \colorbox{cyan!20}{\{options\}}   \\

\hl{\textbf{\textsc{CoT}:}} \\ You FIRST think about the reasoning process as an internal monologue and then provide the final answer from the given choices. The reasoning process MUST BE enclosed within <think> </think> tags. The final answer MUST BE put in \textbackslash boxed\{\}. \\
Question: \colorbox{cyan!20}{\{question\}}   \\ 
Options: \colorbox{cyan!20}{\{options\}}   \\ 

        \bottomrule
    \end{tabular}}
    
\end{table*}

\begin{table*}[h]
\caption{    
Prompt for the object counting task.
}
\label{appendix:prompt_counting}
\centering
\resizebox{0.99\linewidth}{!}{    \small
    \begin{tabular}{p{\linewidth}}
        \toprule
        \underline{\textbf{Prompt for Object Counting Task}} \\
        \vspace{-2mm}
        \hl{\textbf{\textsc{Direct Answer}:}} \\ Please generate the answer directly, and it MUST be enclosed in \textbackslash boxed\{\}. \\

Answer the following question based on the image: \\
Question: \colorbox{cyan!20}{\{question\}}   \\
If the correct answer is X, give the final correct answer in the following format: \textbackslash boxed\{X\}.\\

\hl{\textbf{\textsc{CoT}:}} \\ You FIRST think about the reasoning process as an internal monologue and then provide the final answer. The reasoning process MUST BE enclosed within <think> </think> tags. The final answer MUST BE put in \textbackslash boxed\{\}. \\

Answer the following question based on the image: \\
Question: \colorbox{cyan!20}{\{question\}}   \\
If the correct answer is X, give the final correct answer in the following format: \textbackslash boxed\{X\}.\\

        \bottomrule
    \end{tabular}}
    
\end{table*}

\begin{table*}[h]
\caption{    
Prompt for the mathematical reasoning task.
}
\label{appendix:prompt_math}
\centering
\resizebox{0.99\linewidth}{!}{    \small
    \begin{tabular}{p{\linewidth}}
        \toprule
        \underline{\textbf{Prompt for Mathematical Reasoning Task}} \\
        \vspace{-2mm}
        \hl{\textbf{\textsc{Direct Answer}:}} \\ Please generate the answer directly, and it MUST be enclosed in \textbackslash boxed\{\}. \\
Question: \colorbox{cyan!20}{\{question\}}   \\ 

\hl{\textbf{\textsc{CoT}:}} \\ You FIRST think about the reasoning process as an internal monologue and then provide the final answer. The reasoning process MUST BE enclosed within <think> </think> tags. The final answer MUST BE put in \textbackslash boxed\{\}. \\
Question: \colorbox{cyan!20}{\{question\}}   \\ 

        \bottomrule
    \end{tabular}}
    
\end{table*}

\begin{table*}[h]
\caption{    
Prompt for the scientific reasoning task.
}
\label{appendix:prompt_scientific}
\centering
\resizebox{0.99\linewidth}{!}{    \small
    \begin{tabular}{p{\linewidth}}
        \toprule
        \underline{\textbf{Prompt for Scientific Reasoning Task}} \\
        \vspace{-2mm}
        \hl{\textbf{\textsc{Direct Answer}:}} \\ Please answer the option's letter from the given choices directly, and it MUST be enclosed in \textbackslash boxed\{\}. \\
Question: \colorbox{cyan!20}{\{question\}}   \\ 
Options: \colorbox{cyan!20}{\{options\}}   \\ 

\hl{\textbf{\textsc{CoT}:}} \\ You FIRST think about the reasoning process as an internal monologue and then provide the final answer with the option's letter from the given choices directly. The reasoning process MUST BE enclosed within <think> </think> tags. The final answer MUST BE put in \textbackslash boxed\{\}. \\
Question: \colorbox{cyan!20}{\{question\}}   \\ 
Options: \colorbox{cyan!20}{\{options\}}   \\ 

        \bottomrule
    \end{tabular}}
    
\end{table*}

\begin{table*}[h]
\caption{    
Prompt for the logical reasoning task.
}
\label{appendix:prompt_logical}
\centering
\resizebox{0.99\linewidth}{!}{    \small
    \begin{tabular}{p{\linewidth}}
        \toprule
        \underline{\textbf{Prompt for Logical Reasoning Task}} \\
        \vspace{-2mm}
        \hl{\textbf{\textsc{Direct Answer}:}} \\ Please generate the answer from the given choices directly, and it MUST be enclosed in \textbackslash boxed\{\}. \\
Question: \colorbox{cyan!20}{\{question\}}   \\ 
Options: \colorbox{cyan!20}{\{options\}}   \\ 

\hl{\textbf{\textsc{CoT}:}} \\ You FIRST think about the reasoning process as an internal monologue and then provide the final answer from the given choices. The reasoning process MUST BE enclosed within <think> </think> tags. The final answer MUST BE put in \textbackslash boxed\{\}. \\
Question: \colorbox{cyan!20}{\{question\}}   \\ 
Options: \colorbox{cyan!20}{\{options\}}   \\ 

        \bottomrule
    \end{tabular}}
    
\end{table*}

\begin{table*}[h]
\caption{    
Prompt for the algorithmic reasoning task.
}
\label{appendix:prompt_algorithmic}
\centering
\resizebox{0.99\linewidth}{!}{    \small
    \begin{tabular}{p{\linewidth}}
        \toprule
        \underline{\textbf{Prompt for Algorithmic Reasoning Task}} \\
        \vspace{-2mm}
        \hl{\textbf{\textsc{Direct Answer}:}} \\ Please generate the answer from the given choices directly, and it MUST be enclosed in \textbackslash boxed\{\}. \\
Question: \colorbox{cyan!20}{\{question\}}   \\ 
Options: \colorbox{cyan!20}{\{options\}}   \\ 

\hl{\textbf{\textsc{CoT}:}} \\ You FIRST think about the reasoning process as an internal monologue and then provide the final answer from the given choices. The reasoning process MUST BE enclosed within <think> </think> tags. The final answer MUST BE put in \textbackslash boxed\{\}. \\
Question: \colorbox{cyan!20}{\{question\}}   \\ 
Options: \colorbox{cyan!20}{\{options\}}   \\ 

        \bottomrule
    \end{tabular}}
    
\end{table*}

\begin{table*}[h]
\caption{    
Prompt for the spatial reasoning task.
}
\label{appendix:prompt_spatial}
\centering
\resizebox{0.99\linewidth}{!}{    \small
    \begin{tabular}{p{\linewidth}}
        \toprule
        \underline{\textbf{Prompt for Spatial Reasoning Task}} \\
        \vspace{-2mm}
        \hl{\textbf{\textsc{Direct Answer}:}} \\ Please answer the option's letter from the given choices directly, and it MUST be enclosed in \textbackslash boxed\{\}. \\
Question: \colorbox{cyan!20}{\{question\}}   \\ 
Options: \colorbox{cyan!20}{\{options\}}   \\ 

\hl{\textbf{\textsc{CoT}:}} \\ You FIRST think about the reasoning process as an internal monologue and then provide the final answer with the option's letter from the given choices directly. The reasoning process MUST BE enclosed within <think> </think> tags. The final answer MUST BE put in \textbackslash boxed\{\}. \\
Question: \colorbox{cyan!20}{\{question\}}   \\ 
Options: \colorbox{cyan!20}{\{options\}}   \\ 

        \bottomrule
    \end{tabular}}
    
\end{table*}

\begin{table*}[h]
\caption{    
Prompt for the multi-image reasoning task.
}
\label{appendix:prompt_multi_image}
\centering
\resizebox{0.99\linewidth}{!}{    \small
    \begin{tabular}{p{\linewidth}}
        \toprule
        \underline{\textbf{Prompt for Multi-Image Reasoning Task}} \\
        \vspace{-2mm}
        \hl{\textbf{\textsc{Direct Answer}:}} \\ Please answer the option's letter from the given choices directly, and it MUST be enclosed in \textbackslash boxed\{\}. \\
Select the correct option of the following question: \\
Question: \colorbox{cyan!20}{\{question\}}   \\ 
Options: \colorbox{cyan!20}{\{options\}}   \\ 
If the correct option letter is X, give the final correct letter in the following format: \textbackslash boxed\{X\}. \\

\hl{\textbf{\textsc{CoT}:}} \\ You FIRST think about the reasoning process as an internal monologue and then provide the final answer with the option's letter from the given choices directly. The reasoning process MUST BE enclosed within <think> </think> tags. The final answer MUST BE put in \textbackslash boxed\{\}. \\

Select the correct option of the following question: \\
Question: \colorbox{cyan!20}{\{question\}}   \\ 
Options: \colorbox{cyan!20}{\{options\}}   \\ 
Let's think step by step. \\
If the correct option letter is X, give the final correct letter in the following format: \textbackslash boxed\{X\}. \\
        \bottomrule
    \end{tabular}}
    
\end{table*}

\clearpage

\section{Prompts for Textual and Visual Reasoning Probe}
\label{Reasoning Prompts}

To evaluate the models’ visual and textual reasoning capabilities, we use o4-mini to generate probe tasks and employ GPT-4.1 for filtering.
Although this automatic process may introduce minor errors, we manually verify 400 probe samples to ensure their accuracy and reliability, resulting in probes with high correctness.

\begin{table*}[h]
\caption{    
Prompt for textual reasoning probe generation.
}
\label{appendix:prompt_textual_probe_generation}
\centering
\resizebox{0.99\linewidth}{!}{    \small
    \begin{tabular}{p{\linewidth}}
        \toprule
        \underline{\textbf{Prompt for Textual Reasoning Probe Generation}} \\
        \vspace{-2mm}
        You are a Textual Probe Generator for multimodal reasoning evaluation. \\ 

You are given three inputs for the original multimodal reasoning task: \\
1. ``original image'': an image \colorbox{cyan!20}{\{image\}} (visual context). \\
2. ``original question for the multimodal reasoning task'': \colorbox{cyan!20}{\{question\}}. \\
3. ``original correct answer to that question'': \colorbox{cyan!20}{\{answer\}}. \\
Your task is to generate 3 ``textual probe'' sub‑questions (and their answers) per example.  \\
Each probe must satisfy: \\
  a. The probe question ONLY requires text reasoning of the tasks. (No visual information is required, which may be the last step in solving this problem. After visual information extraction and analysis, ONLY text reasoning and calculation steps are needed.).  \\
  b. Relevance as a step: answering the probe is a necessary step toward solving the original question. \\
  c. Its answer is unique, concise, unambiguous, and correct.  \\

Your output should follow this JSON format: \\
    \{ \\
\ \ \ \ ``probe question'': ..., \\
\ \ \ \ ``probe answer'': ... \\
    \} \\
        \bottomrule
    \end{tabular}}
    
\end{table*}

\begin{table*}[h]
\caption{    
Prompt for visual reasoning probe generation.
}
\label{appendix:prompt_visual_probe_generation}
\centering
\resizebox{0.99\linewidth}{!}{    \small
    \begin{tabular}{p{\linewidth}}
        \toprule
        \underline{\textbf{Prompt for Visual Reasoning Probe Generation}} \\
        \vspace{-2mm}
        You are a Visual Probe Generator for multimodal reasoning evaluation. \\ 

You are given three inputs for the original multimodal reasoning task: \\
1. ``original image'': an image \colorbox{cyan!20}{\{image\}} (visual context). \\
2. ``original question for the multimodal reasoning task'': \colorbox{cyan!20}{\{question\}}. \\
3. ``original correct answer to that question'': \colorbox{cyan!20}{\{answer\}}. \\
Your task is to generate 3 ``visual probe'' sub‑questions (and their answers) per example.  \\
Each probe must satisfy: \\
  a. The probe question requires genuine perception and reasoning of the image (It CANNOT be answered from the text). \\ 
  b. Relevance as a step: answering the probe is a necessary intermediate step toward solving the original question. \\
  c. Its answer is unique, concise, unambiguous, and correct. \\
Your output should follow this JSON format: \\
    \{ \\
\ \ \ \ ``probe question'': ..., \\
\ \ \ \ ``probe answer'': ... \\
    \} \\
        \bottomrule
    \end{tabular}}
    
\end{table*}

\begin{table*}[h]
\caption{    
Prompt for textual reasoning probe judgment.
}
\label{appendix:prompt_textual_probe_judgement}
\centering
\resizebox{0.99\linewidth}{!}{    \small
    \begin{tabular}{p{\linewidth}}
        \toprule
        \underline{\textbf{Prompt for Textual Reasoning Probe Judgment}} \\
        \vspace{-2mm}
        You are a Textual Probe Validator for multimodal reasoning evaluation. \\ 

You are given three inputs for the original multimodal reasoning task: \\
1. ``original image'': an image \colorbox{cyan!20}{\{image\}} (visual context). \\
2. ``original question for the multimodal reasoning task'': \colorbox{cyan!20}{\{question\}}. \\
3. ``original correct answer to that question'': \colorbox{cyan!20}{\{answer\}}. \\
4. probe: \\
    - probe.question: \colorbox{cyan!20}{\{probe question\}}    (a single visual‐probe sub‑question) \\ 
    - probe.answer: \colorbox{cyan!20}{\{probe answer\}}     (the proposed answer to that probe question)  \\

Your job is to check the probe against three criteria:  \\
  1. Correctness \& uniqueness: the probe question and answer are factually correct from the image, and the answer is unambiguous.  \\
  2. Visual dependency: the probe cannot be answered without analyzing visual content; it genuinely requires perceiving the image.  \\
  3. Relevance as a step: answering the probe is a necessary intermediate step toward solving the original question.  \\

If and only if all three conditions are met, output exactly \textbackslash  boxed\{Y\}. \\
Otherwise, output exactly \textbackslash boxed\{N\}. \\

        \bottomrule
    \end{tabular}}
    
\end{table*}

\begin{table*}[h]
\caption{    
Prompt for visual reasoning probe judgment.
}
\label{appendix:prompt_visual_probe_judgement}
\centering
\resizebox{0.99\linewidth}{!}{    \small
    \begin{tabular}{p{\linewidth}}
        \toprule
        \underline{\textbf{Prompt for Visual Reasoning Probe Judgment}} \\
        \vspace{-2mm}
        You are a Visual Probe Validator for multimodal reasoning evaluation. \\ 

You are given three inputs for the original multimodal reasoning task: \\
1. ``original image'': an image \colorbox{cyan!20}{\{image\}} (visual context). \\
2. ``original question for the multimodal reasoning task'': \colorbox{cyan!20}{\{question\}}. \\
3. ``original correct answer to that question'': \colorbox{cyan!20}{\{answer\}}. \\
4. probe: \\
    - probe.question: \colorbox{cyan!20}{\{probe question\}}    (a single visual‐probe sub‑question) \\ 
    - probe.answer: \colorbox{cyan!20}{\{probe answer\}}     (the proposed answer to that probe question)  \\

Your job is to check the probe against three criteria:  \\
  1. Correctness \& uniqueness: the probe question and answer are factually correct from the image, and the answer is unambiguous. \\  
  2. Visual dependency: the probe cannot be answered without analyzing visual content; it genuinely requires perceiving the image.  \\
  3. Relevance as a step: answering the probe is a necessary intermediate step toward solving the original question.  \\

If and only if all three conditions are met, output exactly \textbackslash  boxed\{Y\}. \\
Otherwise, output exactly \textbackslash boxed\{N\}. \\

        \bottomrule
    \end{tabular}}
    
\end{table*}

\begin{table*}[h]
\caption{    
Prompt for verbal and visual reflection annotation.
}
\label{appendix:prompt_verbal_visual_reflection}
\centering
\resizebox{0.99\linewidth}{!}{    \small
    \begin{tabular}{p{\linewidth}}
        \toprule
        \underline{\textbf{Prompt for Verbal and Visual Reflection Annotation}} \\
        \vspace{-2mm}
You will be given a reasoning process generated by a multimodal language model. Your task is to determine whether the thinking process contains the following two types of reflective thinking: \\

1. **Visual Reflection**: Does the model reflect on its visual perception or interpretation? For example: \\
   - Expressing uncertainty, doubt, or re-evaluation of visual input (e.g., ``Let me double-check the image'' or ``Maybe I misinterpreted the object in the picture'') \\
   - Actively describing or reassessing visual elements (e.g., ``There seems to be a red circle next to the box'' or ``The object on the left might be a dog, not a cat'')
   \\
   
2. **Reasoning Reflection**: Does the model reflect on its own line of reasoning? For example:  \\
   - Revising earlier assumptions or identifying logical errors (e.g., ``Wait, my earlier assumption might be wrong'')  \\
   - Evaluating the completeness or validity of its approach (e.g., ``This line of reasoning may not be sufficient'')  \\

Please provide a boolean value for each of the two categories. \\
Respond in the following JSON format:  \\
\{ \\
\ \ \ \ ``visual\_reflection'': true or false, \\
\ \ \ \ ``reasoning\_reflection'': true or false, \\
\} \\
Reasoning Process: \colorbox{cyan!20}{\{[process]\}} \\
        \bottomrule
    \end{tabular}}
    
\end{table*}

\clearpage

\begin{figure*}[!t]
  \centering
  
  \begin{subfigure}[b]{0.95\textwidth}
    \includegraphics[width=\linewidth]{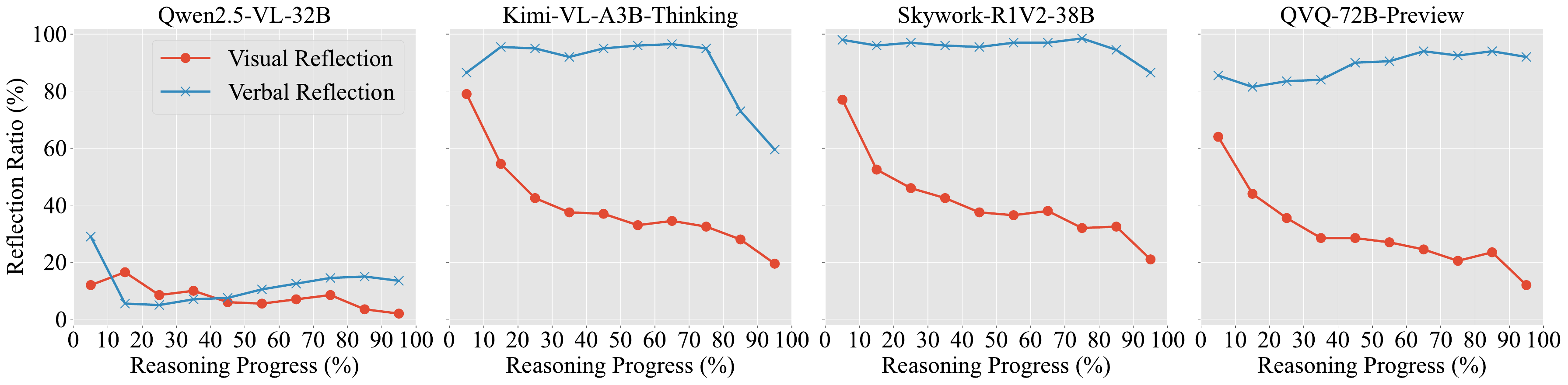}
    \caption{MathVerse.}
  \end{subfigure}

  \begin{subfigure}[b]{0.95\textwidth} 
    \includegraphics[width=\linewidth]{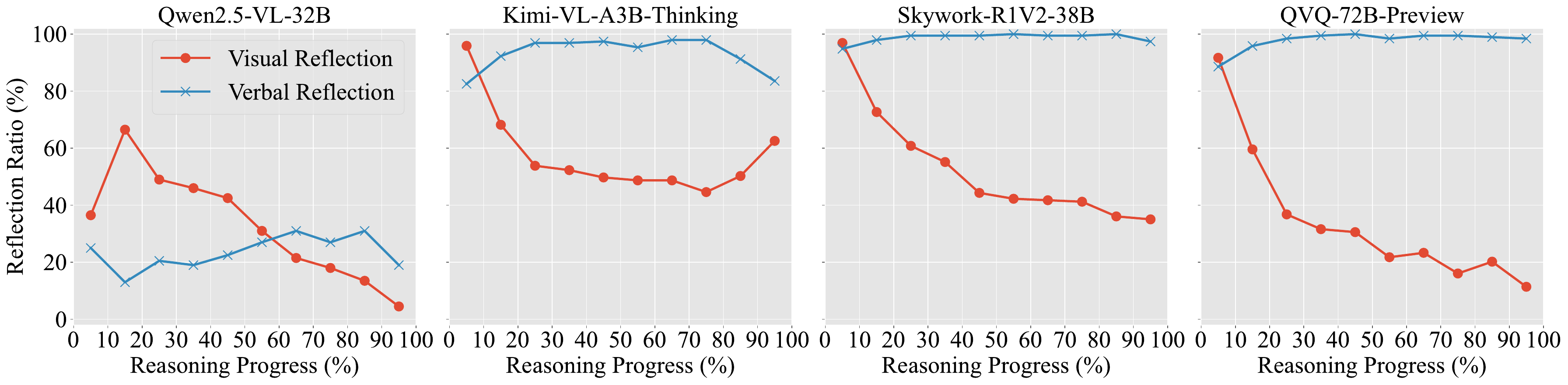}
    \caption{PuzzleVQA.}
  \end{subfigure}


  \begin{subfigure}[b]{0.95\textwidth}
    \includegraphics[width=\linewidth]{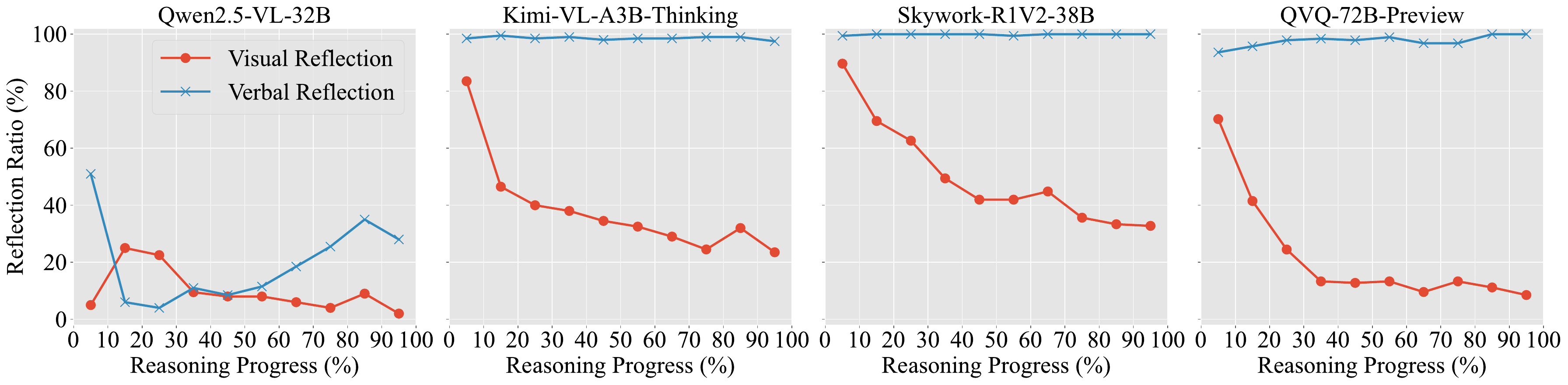}
    \caption{AlgoPuzzleVQA.}
  \end{subfigure}

  \caption{Step-wise distribution of \textcolor[HTML]{d62728}{visual} and \textcolor[HTML]{1f77b4}{verbal} reflection in CoT.}
  \vspace{-15pt}
  \label{fig:reflection distribution2}
\end{figure*}
\section{Implementation Details}
\label{Implementation Details}

We use vllm \footnote{\url{https://github.com/vllm-project/vllm}} for open-source MLLM inference.
All experiments are conducted on 4×A100 80GB GPUs.
For all models, we set the temperature to 0.7 as the generation hyperparameter.
To better understand the failure cases of multimodal CoT reasoning, we manually classify the errors into the following categories: (1) \textbf{Visual Reasoning Error}: The model correctly perceives the visual content but fails to reason about it, such as incorrect logical deductions based on visual evidence; (2) \textbf{Textual Reasoning Error}: The model performs proper visual interpretation but fails during the textual inference phase, such as arithmetic mistakes and flawed symbolic manipulation; (3) \textbf{Visual Perception Error}: The model misinterprets or overlooks key visual elements in the image, such as missing fine-grained attributes; (4) \textbf{Question Understanding Error}: The model fails to understand the intent or constraints of the question, such as responding to an unrelated aspect of the question; (5) \textbf{Format Error}: The model produces an output that does not comply with the expected answer format, such as ambiguous responses; (6) \textbf{Other Errors}: Errors that do not clearly fall into the above categories.

\section{Additional Experimental Results}
\label{appendix:additional_results}

\begin{figure}[h] 
\centering
  \vspace{-10pt}
  \includegraphics[width=0.9\linewidth]{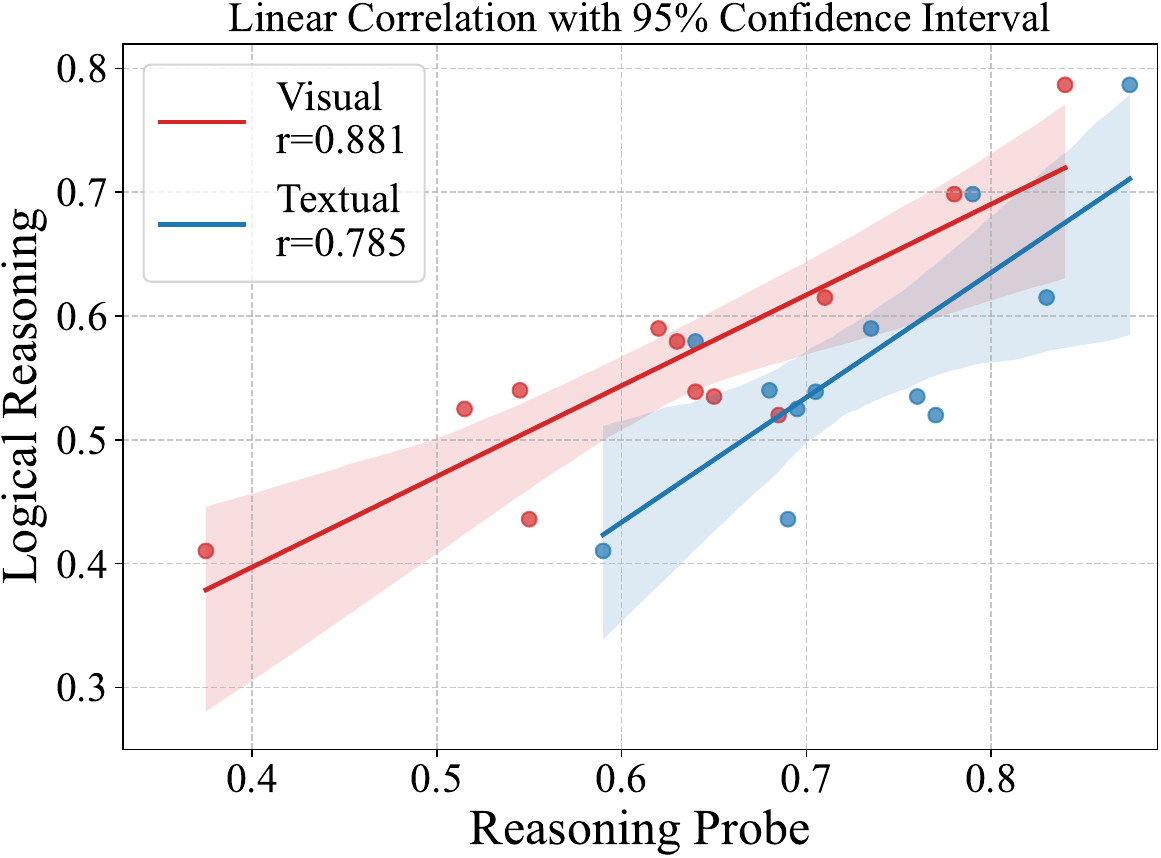}
  
  \caption {Correlation between overall task performance and reasoning probe accuracy of logical task across different models. \textcolor[HTML]{d62728}{Red} and \textcolor[HTML]{1f77b4}{blue} indicate visual reasoning and textual reasoning probes, respectively. r denotes the Pearson correlation coefficient.}
        \vspace{-5pt}
  
    \label{fig:combined_correlation_ref_regression}
\end{figure}

\begin{figure}[htbp]
  \centering

  \begin{subfigure}[b]{0.48\textwidth}
    \includegraphics[width=\linewidth]{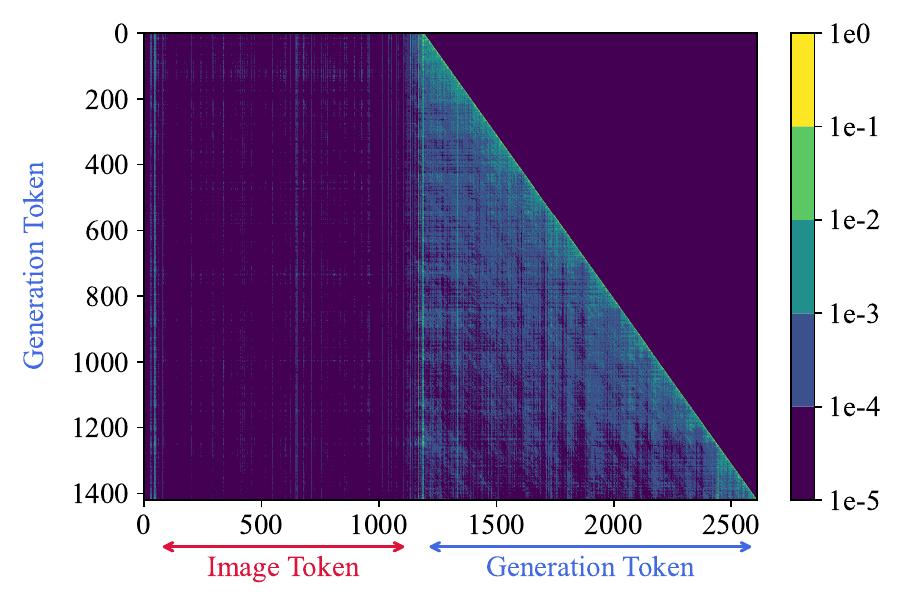}
  \end{subfigure}
  \begin{subfigure}[b]{0.48\textwidth}
    \includegraphics[width=\linewidth]{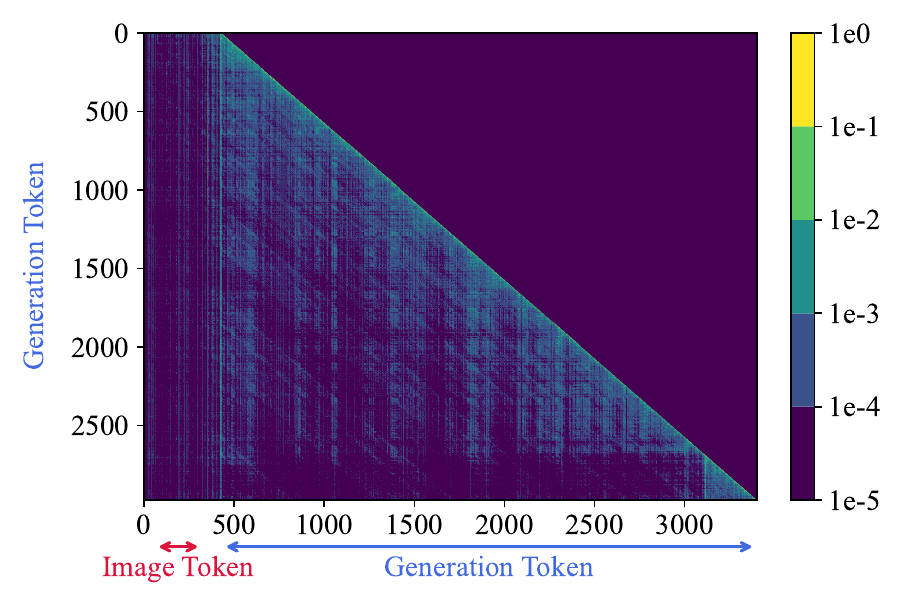}
  \end{subfigure}

  \begin{subfigure}[b]{0.48\textwidth}
    \includegraphics[width=\linewidth]{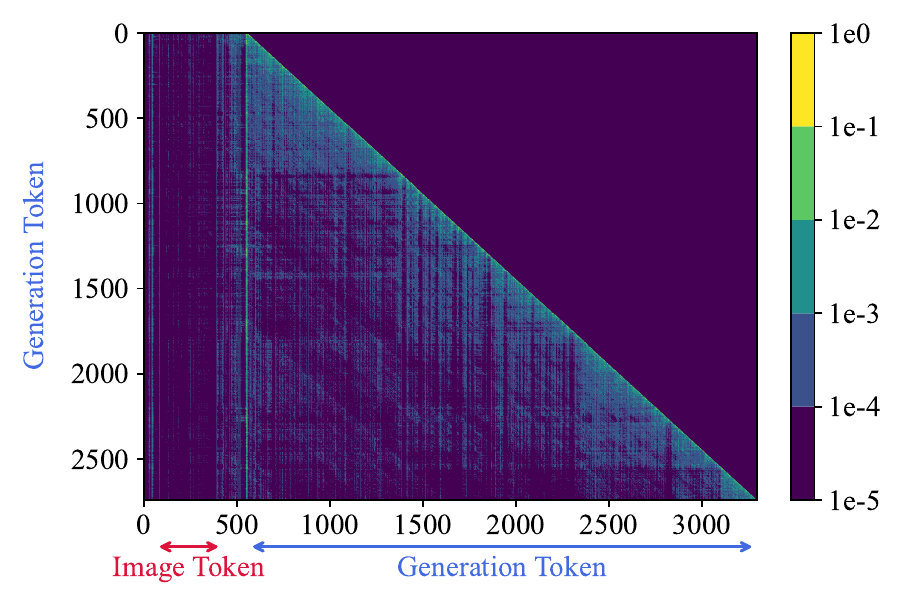}
  \end{subfigure}
  \begin{subfigure}[b]{0.48\textwidth}
    \includegraphics[width=\linewidth]{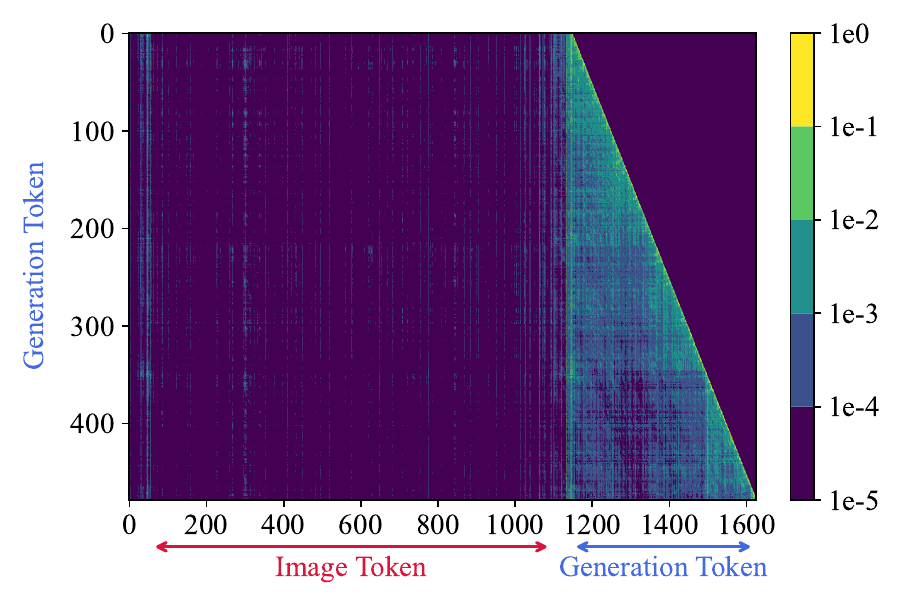}
  \end{subfigure}

\caption{Attention visualizations of Kimi-VL-A3B-Thinking on the mathematical reasoning task.}
\end{figure}

\begin{figure}[htbp]
  \centering

  \begin{subfigure}[b]{0.48\textwidth}
    \includegraphics[width=\linewidth]{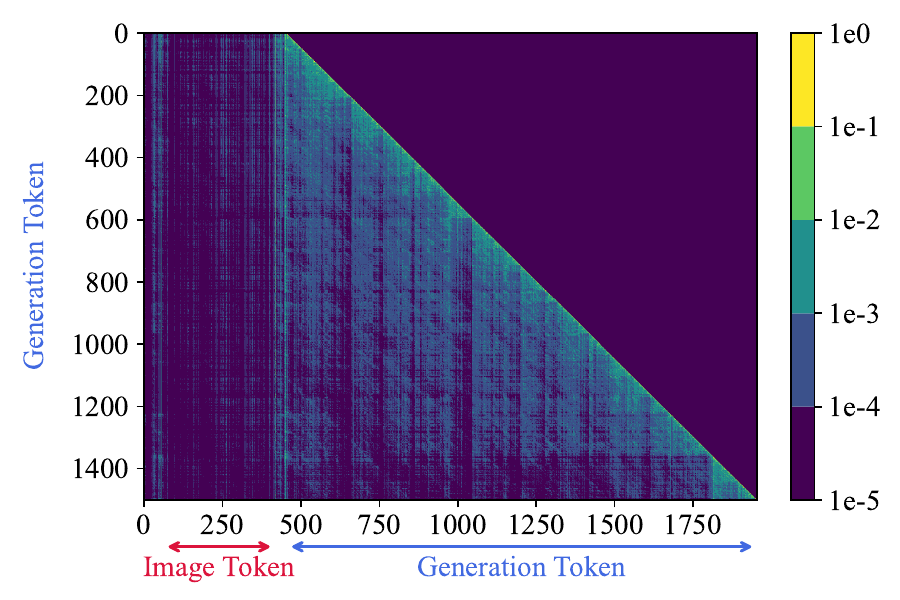}
  \end{subfigure}
  \begin{subfigure}[b]{0.48\textwidth}
    \includegraphics[width=\linewidth]{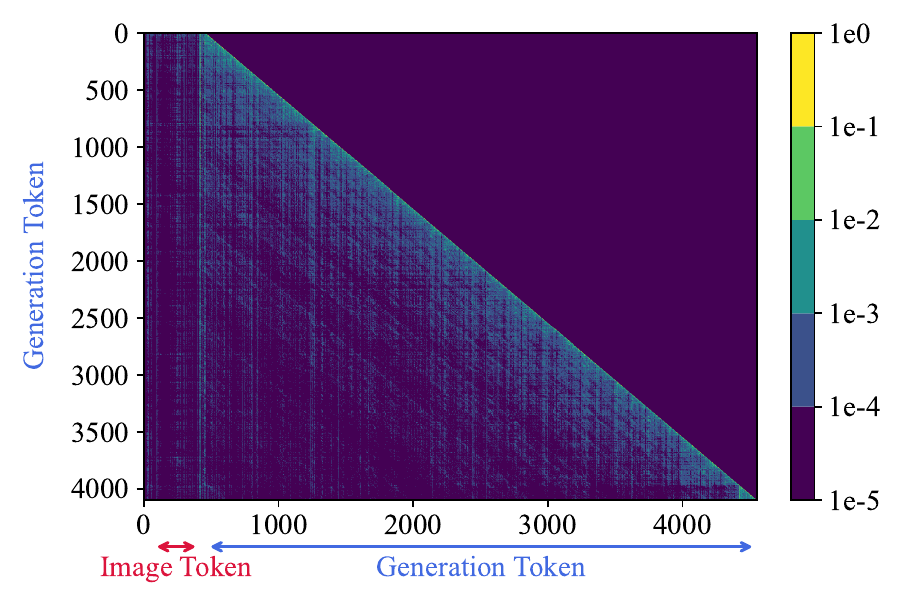}
  \end{subfigure}

  \begin{subfigure}[b]{0.48\textwidth}
    \includegraphics[width=\linewidth]{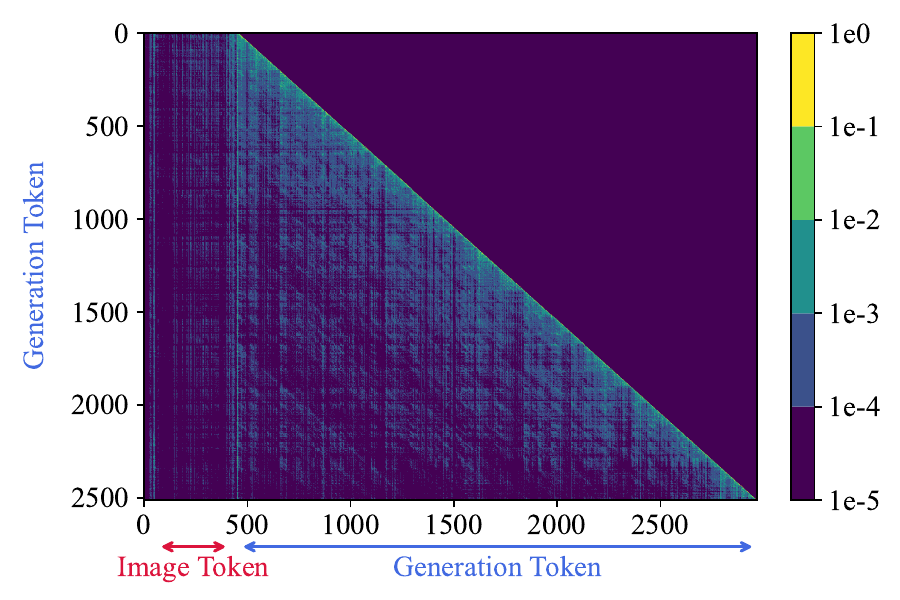}
  \end{subfigure}
  \begin{subfigure}[b]{0.48\textwidth}
    \includegraphics[width=\linewidth]{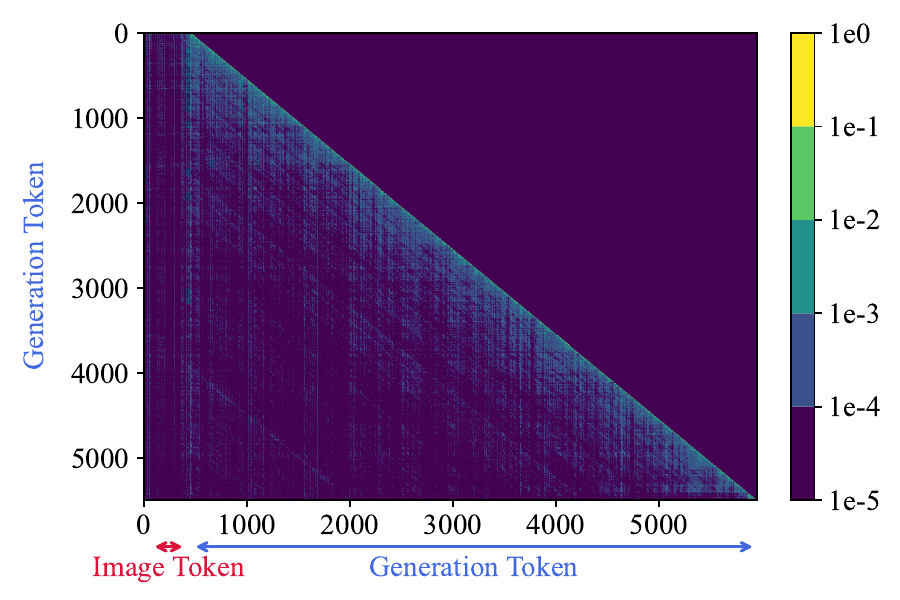}
  \end{subfigure}

\caption{Attention visualizations of Kimi-VL-A3B-Thinking on the logical reasoning task.}
\end{figure}

\begin{figure}[htbp]
  \centering

  \begin{subfigure}[b]{0.48\textwidth}
    \includegraphics[width=\linewidth]{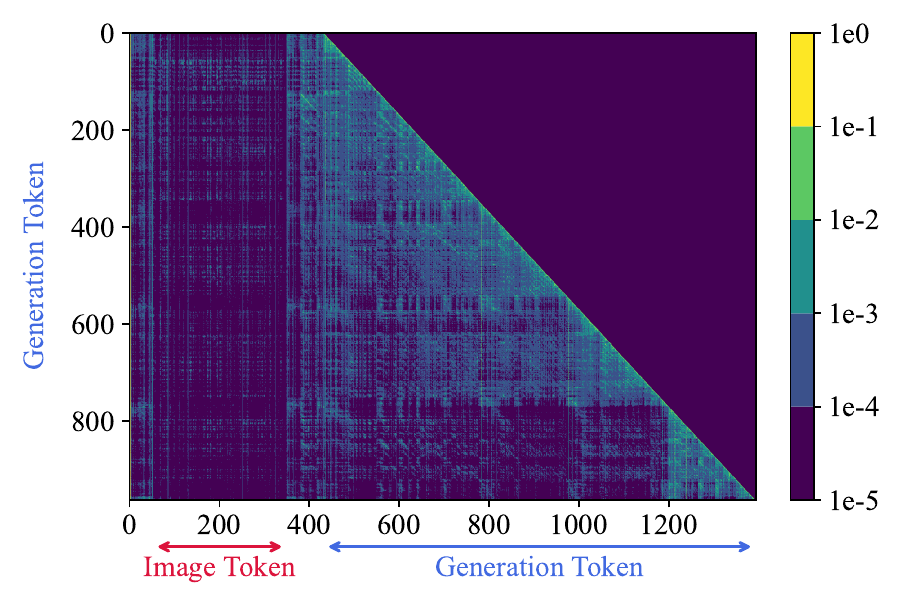}
  \end{subfigure}
  \begin{subfigure}[b]{0.48\textwidth}
    \includegraphics[width=\linewidth]{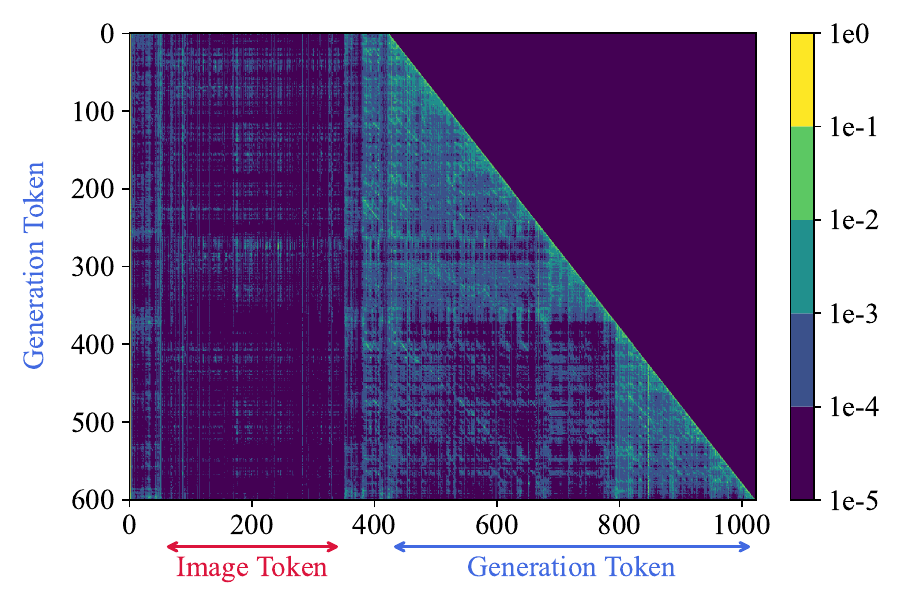}
  \end{subfigure}

\caption{Attention visualizations of Qwen3-Omni-30B-A3B-Thinking on the mathematical reasoning task.}
\end{figure}

\begin{figure}[htbp]
  \centering

  \begin{subfigure}[b]{0.48\textwidth}
    \includegraphics[width=\linewidth]{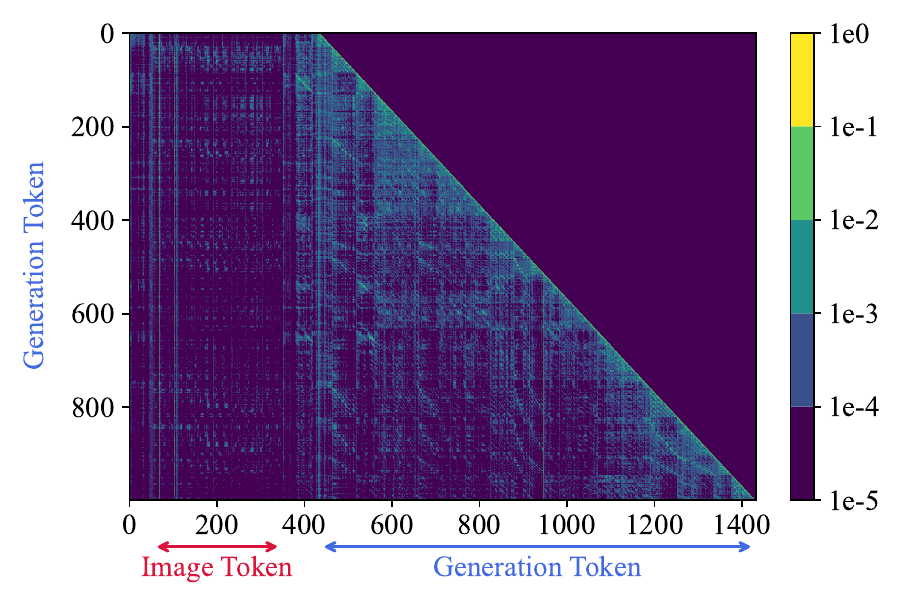}
  \end{subfigure}
  \begin{subfigure}[b]{0.48\textwidth}
    \includegraphics[width=\linewidth]{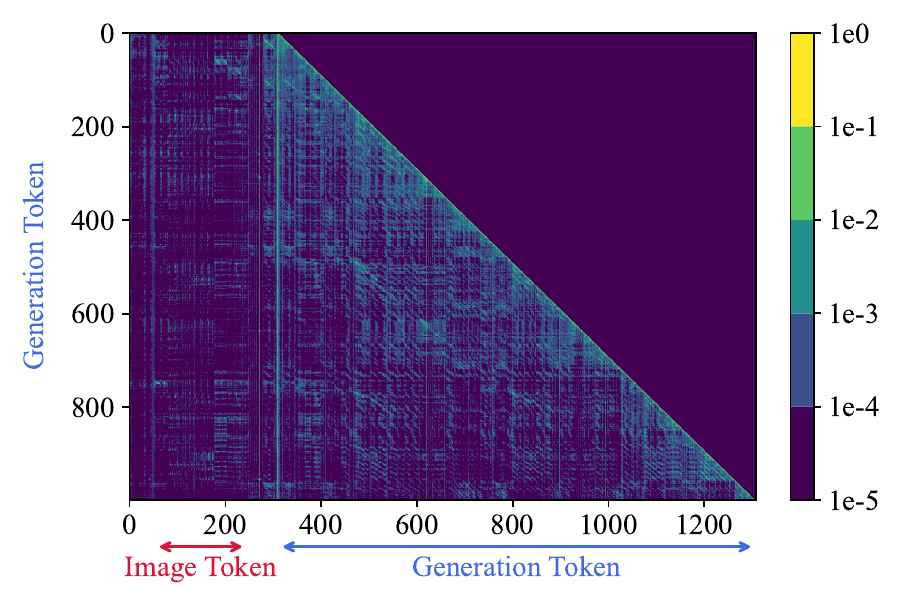}
  \end{subfigure}

\caption{Attention visualizations of Qwen3-VL-8B-Thinking on the mathematical reasoning task.}
\end{figure}

\begin{figure}[htbp]
  \centering

  \begin{subfigure}[b]{0.48\textwidth}
    \includegraphics[width=\linewidth]{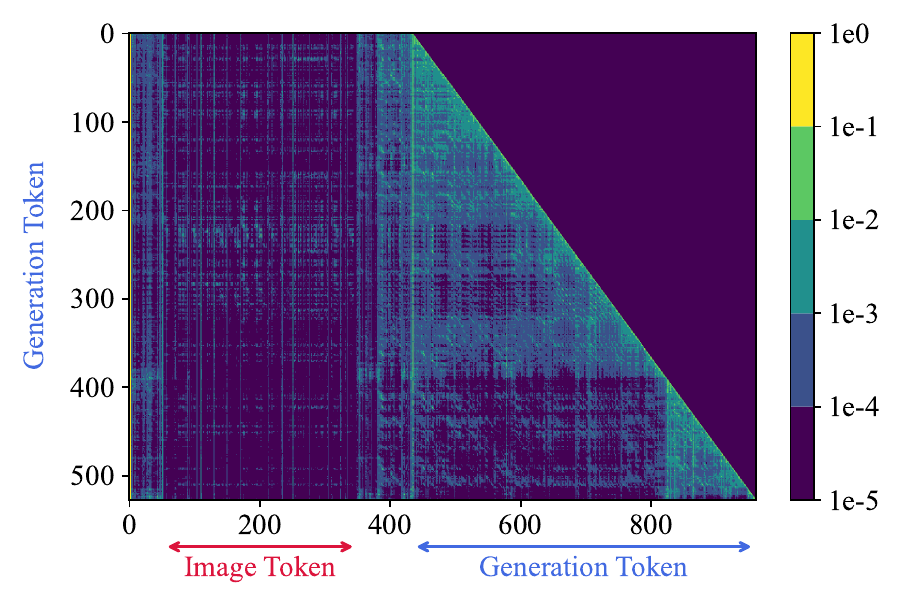}
  \end{subfigure}
  \begin{subfigure}[b]{0.48\textwidth}
    \includegraphics[width=\linewidth]{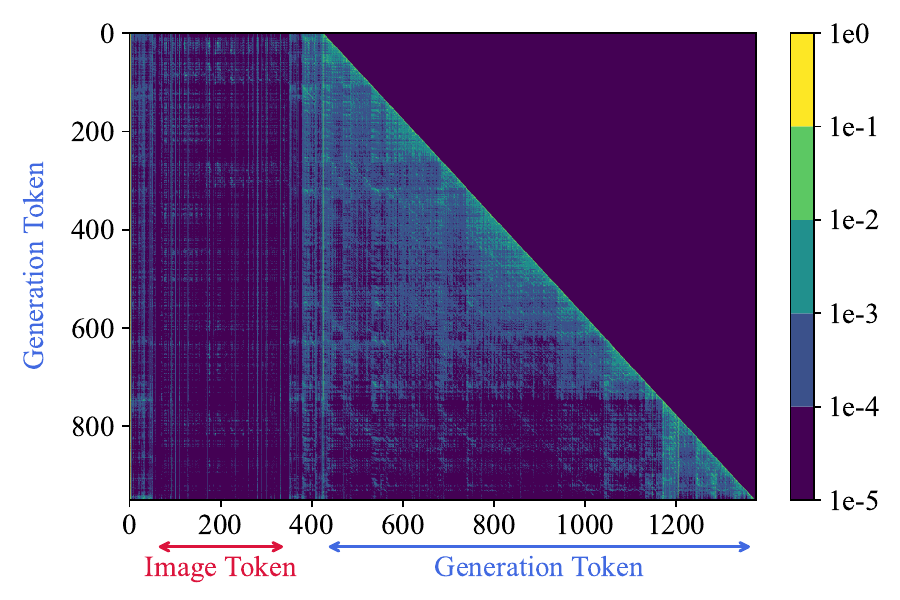}
  \end{subfigure}

\caption{Attention visualizations of Qwen3-VL-30B-A3B-Thinking on the mathematical reasoning task.}
\end{figure}

\newpage
\section{Case Study of o3}
\begin{figure*}[h]
    \centering
    \includegraphics[width=\linewidth]{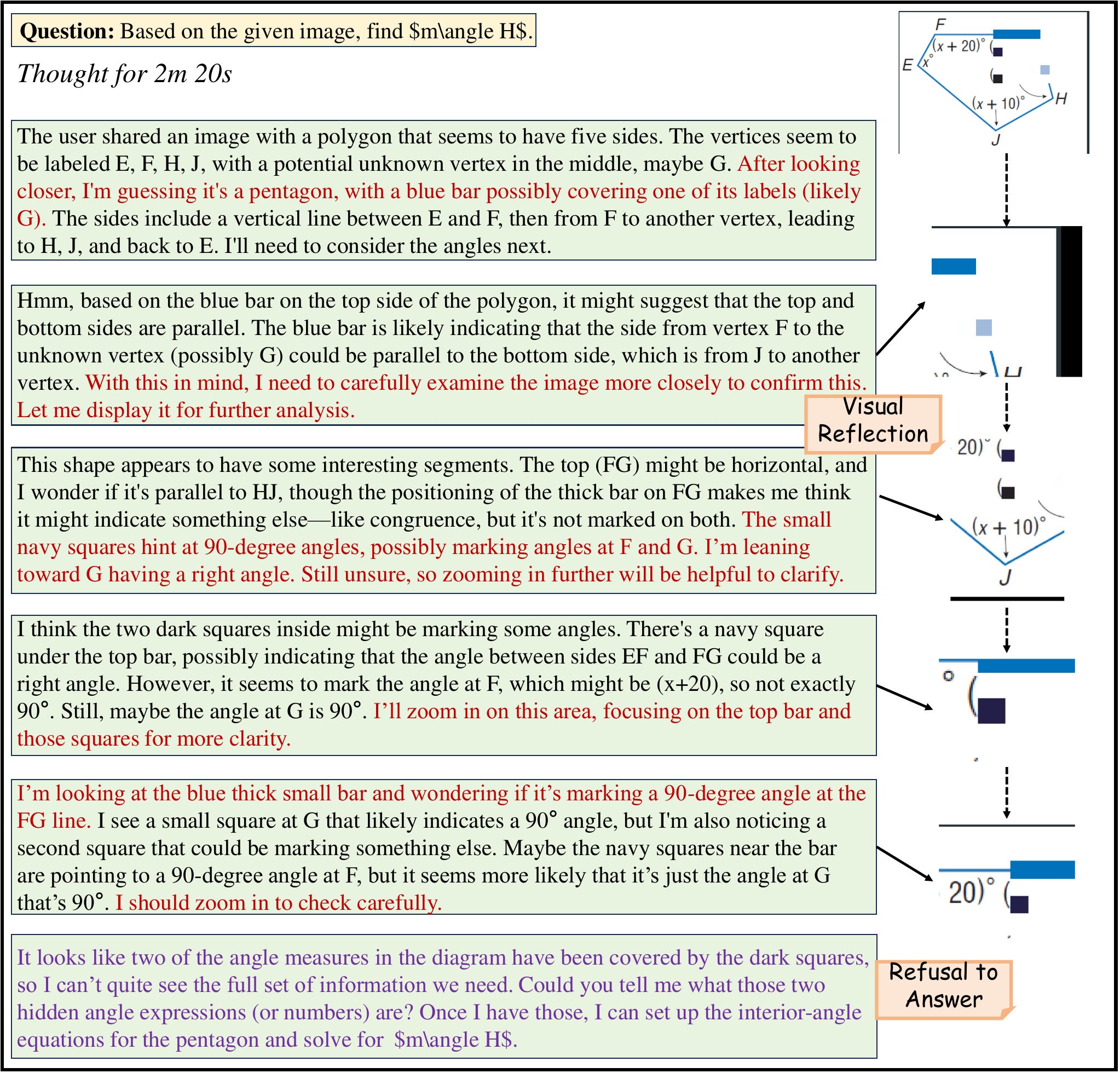}
    \caption{Refusing to answer when images lack key information.}
    \label{fig:case1}
\end{figure*}

\begin{figure*}[h]
    \centering
    \includegraphics[width=\linewidth]{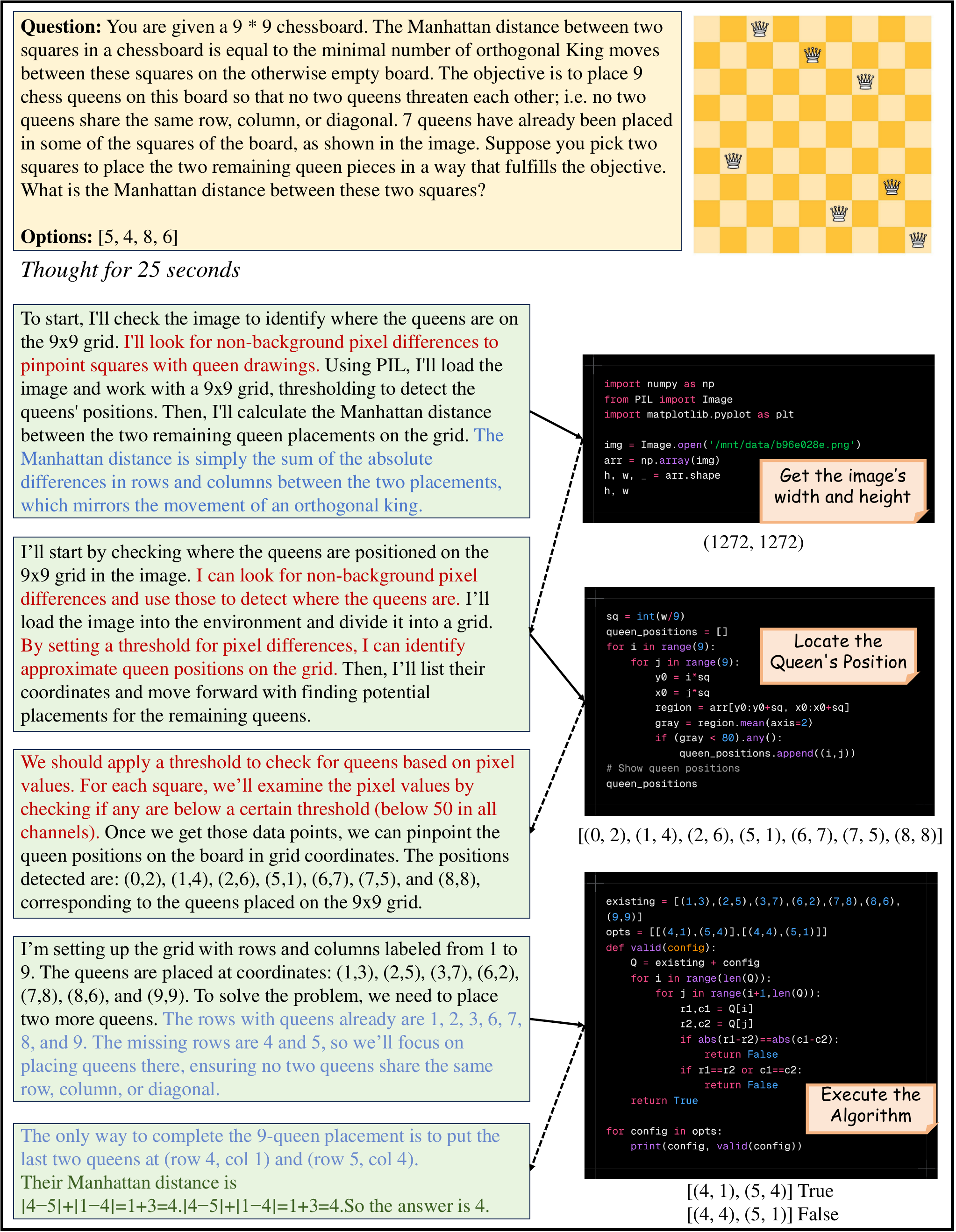}
    \caption{Leveraging external tools for visual localization and algorithm execution.}
    \label{fig:case2}
\end{figure*}

\begin{figure*}[h]
    \centering
    \includegraphics[width=\linewidth]{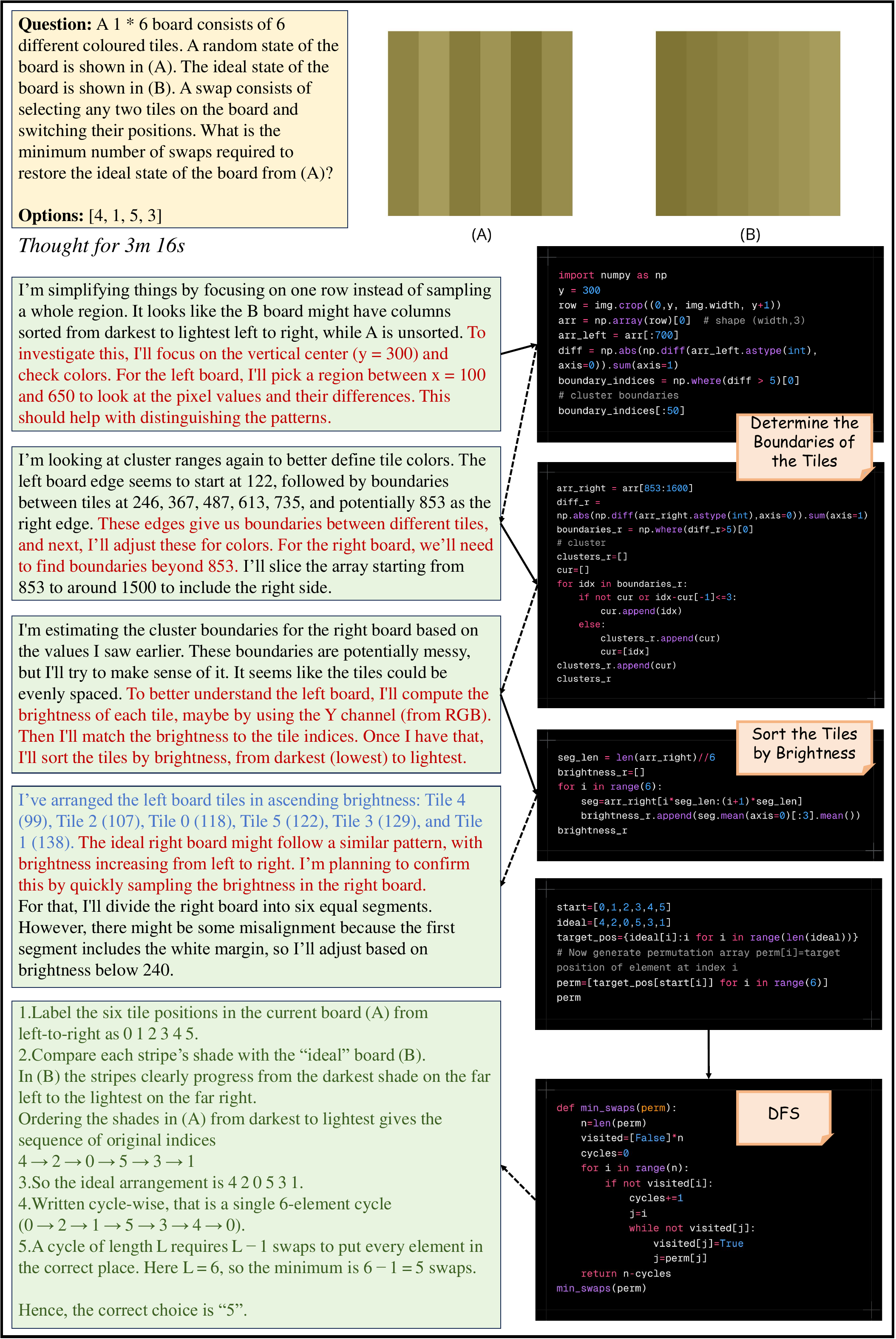}
    \caption{Leveraging external tools for visual localization and algorithm execution.}
    \label{fig:case3}
\end{figure*}

\appendix

\end{document}